\documentclass{article} 
\usepackage{iclr2026_conference,times}

\usepackage{amsmath,amsfonts,bm}

\def\eqref#1{equation~\ref{#1}}

\def\1{\bm{1}}

\DeclareMathAlphabet{\mathsfit}{\encodingdefault}{\sfdefault}{m}{sl}
\SetMathAlphabet{\mathsfit}{bold}{\encodingdefault}{\sfdefault}{bx}{n}

\usepackage{hyperref}
\usepackage{url}
\usepackage{times}
\usepackage{enumitem}
\usepackage[table]{xcolor}
\usepackage{xcolor}
\usepackage{amsmath}
\usepackage{amssymb}
\usepackage{multirow}
\usepackage{multicol}
\usepackage{algorithm}
\usepackage{subcaption}
\usepackage{algorithmic}
\usepackage{graphicx}
\usepackage{tcolorbox}
\newcommand{\cross}{\textcolor{red}{$\times$}}
\newcommand{\tick}{\textcolor{green}{$\checkmark$}}
\newcommand{\approxsym}{\textcolor{gray}{$\approx$}}

\definecolor{first}{HTML}{FC8D58}
\definecolor{second}{HTML}{fdd49f}
\definecolor{third}{HTML}{fff6ee}
\newcommand{\colorsquare}[1]{\colorbox{#1}{\hspace{6pt}\vphantom{A}}}

\usepackage{dblfloatfix}
\usepackage{booktabs}
\newcommand{\minisection}[1]{\vspace{0.02in}\noindent{\bf #1}}

\title{GenColorBench: A Color Evaluation Benchmark for Text-to-Image Generation Models}

\author{
Muhammad Atif Butt$^{1,2}$, Alexandra Gomez-Villa$^{1,2}$, Tao Wu$^{1,2}$, Javier Vazquez-Corral$^{1,2}$, \\
\textbf{Joost Van De Weijer$^{1,2}$, \& Kai Wang$^{1,3,4}$} \\
$^{1}$ Computer Vision Center, Spain \\
$^{2}$ Computer Sciences Department, Universitat Autònoma de Barcelona, Spain \\
$^{3}$ Program of Computer Science, City University of Hong Kong (Dongguan) \\
$^{4}$ City University of Hong Kong
}

\iclrfinalcopy 
\begin{document}

\maketitle

\begin{abstract}
Recent years have seen impressive advances in text-to-image generation, with image generative or unified models, generating high-quality images from text. Yet these models still struggle with fine-grained color controllability, often failing to accurately match colors specified in text prompts. While existing benchmarks evaluate compositional reasoning and prompt adherence, none systematically assess the color precision. Color is fundamental to human visual perception and communication, critical for applications from art to design workflows requiring brand consistency. However, current benchmarks either neglect color or rely on coarse assessments, missing key capabilities like interpreting RGB values or aligning with human expectations. To this end, we propose GenColorBench, the first comprehensive benchmark for T2I color generation, grounded in color systems like ISCC-NBS and CSS3/X11, including numerical colors which are absent elsewhere. With 44K color-focused prompts covering 400+ colors, it reveals models’ true capabilities via perceptual and automated assessments. Evaluations of popular T2I models using GenColorBench show performance variations, highlighting which color conventions models understand best and identifying failure modes. Our GenColorBench assessments will allow to guide improvements in precise color generation. The benchmark will be made public upon acceptance. 
\end{abstract}

\section{Introduction}
Text-to-image (T2I) generation has witnessed remarkable progress in recent years, with state-of-the-art models like Stable Diffusion~\citep{Rombach_2022_CVPR_stablediffusion} and FLUX~\citep{flux2024} demonstrating unprecedented capabilities in generating high-quality, photorealistic images from text prompts. These advances have enabled diverse applications ranging from creative content generation to automated design workflows. However, despite their impressive overall performance, T2I models still struggle with fine-grained controllability, particularly in generating images that precisely match specific visual attributes described in text prompts~\citep{chefer2023attendandexcite,ge2023richtext}. 
While numerous benchmarks, discussed in Table~\ref{tab1:existing_benchmarks}, have been proposed to evaluate various aspects of T2I model performance—including compositional reasoning~\citep{huang2025t2i,ghosh2023geneval}, prompt adherence~\citep{hu2024ella}, and faithfulness~\citep{hu2023tifa}—none systematically evaluates the critical ability to generate precise colors as specified in text prompts.

Color represents a fundamental dimension of human visual perception and serves as a primary channel for human communication about objects and scenes, with color categories forming a universal basis for describing and distinguishing visual phenomena across cultures~\citep{berlin1991basic,witzel2018color}. This perceptual importance translates directly into practical applications where accurate color generation is essential—from multimedia applications and artistic creation to design workflows requiring brand consistency, aesthetic control and faithful reproduction of real-world scenes. However, existing T2I evaluation benchmarks critically underestimate this importance by either neglecting color evaluation entirely or reducing it to coarse categorical assessments that fail to capture their real color capabilities. Current benchmarks do not assess whether models generate colors that maintain color consistency across different contexts, or produce colors that align with human memory and expectations for familiar objects.

To address this, we propose GenColorBench, the first comprehensive benchmark designed to systematically evaluate the color generation capabilities of T2I models. Unlike existing benchmarks that rely on coarse categorical assessments, our benchmark is grounded in established color naming systems, including the ISCC-NBS, and CSS3/X11, and uniquely incorporates evaluation of numerical color specifications (RGB values and hex codes) that are completely absent from existing benchmarks. With over 44K+ prompts specifically designed for color evaluation covering over 400+ colors, GenColorBench provides both the scale and specificity necessary to reveal models' true color generation capabilities through both perceptual color evaluation and automated assessment methods.

We conduct extensive evaluations of several popular image generation models and unified models using GenColorBench, revealing significant variations in color generation capabilities across different models and color specification methods. Our analysis provides insights into which color naming conventions and numerical representations are most effectively understood by current models, and identifies common failure modes in color generation tasks. The main contributions of this work are threefold: (i) We introduce GenColorBench, a large-scale benchmark containing over 44,464 prompts covering 400+ colors specifically designed to evaluate the capabilities of T2I models across five distinct color generation tasks; (ii) We provide comprehensive evaluations of state-of-the-art T2I models, analyzing their performance on precise color generation and identifying key limitations; (iii) We establish baseline performance metrics and evaluation protocols that can guide future research in improving color controllability in generative models.

\begin{tcolorbox}[colback=gray!10,colframe=gray!50,title=\textbf{MAJOR FINDINGS OF OUR EVALUATION}]
	\begin{itemize}
    \item Current models exhibit notable shortcomings in adhering to precise color specifications, underscoring the urgent need for enhanced color controllability (Table \ref{tab3:results}).
    \item Model performance is tightly linked to category semantics. Categories with strong color associations (e.g. Fruits and Vegetables---yellow bananas, green grass) pose greater challenges (Fig.~\ref{fig:cat_accuracy}, Fig.~\ref{fig:bias_analysis}).
     \item Models are better at understanding basic colors (yellow, pink, blue), while they struggle more with intermediate colors (Fig.~\ref{fig:color_comparisons}(Left)). Similarly, models favor ``light'' and ``dark'' modifiers over more nuanced ones like ``-ish'', suggesting a limited grasp of subtle color variations (Fig.~\ref{fig:color_comparisons}(Right)).



   \item Vision-language models fall short as reliable tools for color evaluation (Table \ref{tab:vqa_results}).
 	\end{itemize}
\end{tcolorbox}

\begin{table*}[t]
\centering
\resizebox{\linewidth}{!}{
    \begin{tabular}{lllcccccl}
    \hline
    & & &   \multicolumn{5}{c}{Color Evaluation Tasks}  &  \\
    \cline{4-8}
    Benchmark   &  Scale  &  Focus & CNA  &   MCC  &   COA &   NCU    &   ICA &  Color Evaluation Methods\\
    \hline
    GenEval~\citep{ghosh2023geneval}          &  553  &   Compositionality   & \tick  & \tick  & \approxsym  & \cross  & \cross  &  Mask2Former + CLIP ViT-L/14 \\
    T2I-CompBench++~\citep{huang2025t2i}      & 6000  & Compositionality     & \tick  & \approxsym  & \cross  & \cross  & \cross &  BLIP-VQA \\
    DPG-Bench~\citep{hu2024ella}              & 1065  & Prompt Adherence     & \tick  &  \tick & \cross  &  \cross & \cross  &  mPLUG-large VQA \\
    TIFA~\citep{hu2023tifa}                   & 1000  & Faithfulness         & \tick  &  \tick & \cross  &  \cross & \cross  &  mPLUG-large VQA \\
    Commonsense-T2I~\citep{fu2024commonsense} & 1000+  & Reasoning           & \approxsym  &  \approxsym  & \cross  &  \cross & \cross  &  self-proposed (accuracy)\\
    Winoground-T2I~\citep{zhu2023contrastive} & 11,000  & Compositionality   & \tick & \tick & \cross &  \cross & \cross  &  Human Rating + DSG-VQA \\
    Wise~\citep{niu2025wise}                  & 1000  & Reasoning            & \approxsym & \approxsym  & \cross  &  \cross & \cross  &  WiScore, Aesthetic Quality \\
    MMMG~\citep{luo2025mmmg}                  &  4456 &   Disciplinary Knowledge  &  \tick & \tick  & \tick &  \cross & \cross  & GPT/Gemini/QWEN VQA  \\
    Partiprompt~\citep{yu2022scaling}         & 1600  &  Compositionality    & \tick  & \tick  & \cross  &  \cross & \cross  & FID \\
    OneIG-Bench~\citep{chang2025oneig}        & 2440  & Compositionality     & \cross & \cross & \cross  &  \cross & \cross  &  FID \\
    DrawBench~\citep{saharia2022photorealistic} & 200  & Compositionality    &  \tick &  \tick & \cross  &  \cross & \cross  &  Human Rating \\
    EvalAlign~\citep{tan2024evalalign}        & 3000   & Compositionality    & \tick  &  \tick &  \cross &  \cross & \cross  & MLLM-VQA  \\
    Evalmuse~\citep{han2024evalmuse}          & 4000  &  Compositionality    & \tick & \tick & \cross &  \cross & \cross  &  FGA-BLIP2, PN-VQA  \\
    \textbf{GenColorBench (Ours)} &  44,464 & Color Understanding  & \tick  &  \tick & \tick  & \tick & \tick &  VQA + Color Metrics \\
    \textbf{GenColorBench-Mini (Ours)} &  $<$ 10K & Color Understanding  & \tick & \tick & \tick  &  \tick & \tick  &  VQA + Color Metrics \\
    \hline
    \end{tabular}
}
\caption{Overview of existing T2I evaluation benchmarks. Abbreviations for color evaluation tasks: CN = Color Name Understanding, MC = Multi-Color Composition, CO = Color–Object Association, NCU = Numeric Color Understanding, ICA = Implicit Color Association. \textit{While these benchmarks are widely adopted for assessing various aspects of T2I generation—such as compositionality, prompt adherence, and reasoning—they lack comprehensive coverage of key color understanding and evaluation tasks. \textbf{GenColorBench} is specifically designed to fill this gap by supporting a broad spectrum of color-related tasks.} (\tick: covered, \cross: not covered, \approxsym: partially covered)}
\label{tab1:existing_benchmarks}
\end{table*}

\vspace{-3mm}
\section{Related Work}
\minisection{T2I Diffusion Models.}
T2I generation has advanced rapidly in recent years. T2I diffusion models~\citep{ho2020ddpm,gu2022vector} emerged as more efficient models surpassing GANs~\citep{goodfellow2020generative}, VAEs~\citep{kingma2013vae}, autoregressive~\citep{esser2021taming} and flow-based~\citep{dinh2014nice,dinh2016density} models in T2I generation. 
Diffusion models are probabilistic generative models aiming to learn data distribution through denoising from Gaussian distribution. These models allow multi-modal conditioning~\citep{song2021ddim},~\citep{meng2022sdedit},~\citep{nichol2021glide} to improve controllability. With recent scaling up the scale of diffusion models, SD3~\citep{esser2024scaling_sd3} and FLUX~\citep{flux2024} have been state-of-the-art T2I models while largely surpassing the previous representatives~\citep{ramesh2022dalle2,chen2023pixart}.

\minisection{Unified Models.}
Recent years have seen major progress in multimodal understanding and image generation models. Yet, these fields have advanced along separate paths, forming distinct architectural paradigms. Autoregressive architectures dominate large language models such as LLaMa~\citep{touvron2023llama}, Qwen~\citep{team2024qwen2}, and multimodal models like LLaVa~\citep{liu2023visual_llava}, Qwen-VL~\citep{qwenvl2024}.
Autoregressive-based architectures have established dominance in large language models such as LLaMa~\citep{touvron2023llama}, Qwen~\citep{team2024qwen2}, etc, as well as in multimodal understanding models including LLaVa~\citep{liu2023visual_llava} and Qwen-VL~\citep{qwenvl2024}.
Diffusion models, such as Stable Diffusion~\citep{podell2023sdxl} and FLUX~\citep{flux2024}, have become central to image generation, producing high-fidelity, prompt-aligned images. More recently, unified frameworks like GPT-4o aim to handle multimodal inputs and outputs in a single mechanism.
Unified models fall into three types: diffusion-based, autoregressive (AR), and fused AR/diffusion. Pure diffusion-based MLLMs, such as MMaDA~\citep{yang2025mmada} and Dual-Diffusion, use dual-branch diffusion for joint text–image generation.
However, unified models based on naive autoregressive (AR) dominate this research landscape, with representative contributions including SEED series~\citep{ge2023making_seedllama}, Emu series~\citep{sun2024generative_emu2}, Janus series~\citep{wu2025janus,chen2025janus_pro}, etc.
Recently, fused AR–diffusion models have emerged for unified vision–language generation, exemplified by Show-o~\citep{xie2024show_o} and BAGEL~\citep{deng2025emerging_bagel}.


\minisection{Color Control in T2I diffusion models.}
With the advancements in generation and unified models,
various text-guided image editing approaches~\citep{hertz2023delta_DDS,meng2022sdedit,mokady2022null} have been developed to enable controllable modifications. 
For instance, methods like Imagic~\citep{kawar2022imagic} and P2P~\citep{hertz2022prompt} leverage Stable Diffusion (SD) models for structure-preserving edits.
And the unified models~\citep{deng2025emerging_bagel,wu2025omnigen2} integrate such editing power by large-scale pretraining with huge paired datasets.
Another technique stream which can also achieve controllable generation is transfer learning for T2I models~\citep{ruiz2022dreambooth,kumari2022customdiffusion}.
It aims at adapting a given model to a \textit{new concept} by given images from the users and bind the new concept with a unique token. As a result, the adaptation model can generate various renditions for the new concept guided by text prompts. 
However, all these existing techniques struggle to achieve fine-grained control over color attributes in image editing and generation tasks. Only a limited number of works~\citep{butt2025colorpeel,ge2023richtext} have begun addressing the challenge of precise color generation. 
To facilitate the evaluation and development of precise color generation capabilities of future models, we build the first color benchmark in this paper.

\minisection{T2I Evaluation.}
A variety of benchmarks have been developed to evaluate text-to-image models, each tailored to specific aspects of generative performance, as listed in Table~\ref{tab1:existing_benchmarks}. GenEval~\citep{ghosh2023geneval} introduces object detectors to enable fine-grained, object-level evaluation, thereby addressing the limitations of holistic metrics.
T2I-CompBench~\citep{huang2025t2i} elevates compositional complexity by constructing prompts that integrate attributes, relational cues, numeracy, and complex scene descriptions. 
DPG-Bench~\citep{hu2024ella} focuses on assessing models’ instruction-following proficiency, leveraging text-rich prompts to gauge their fidelity to detailed directives. 
Furthermore, Commonsense-T2I~\citep{fu2024commonsense} employs adversarial prompts to probe models’ capabilities in visual reasoning.
Winoground-T2I~\citep{zhu2023contrastive} evaluates compositional generalization by leveraging contrastive sentence pairs.
More recently, WISE~\citep{niu2025wise} and MMMG~\citep{luo2025mmmg} benchmarks emphasize world knowledge-based evaluation, spanning cultural, scientific, and temporal domains to gauge models’ alignment with broader understanding.
However, these existing benchmarks are primarily designed to evaluate the general generative capabilities of diverse image generators, with none specifically focusing on the task of color generation. 
A concurrent work, ColorBench \citep{liang2025colorbench}, introduced the first color evaluation benchmark for vision-language models (VLMs). Its focus lies on image color understanding tasks, including color perception, color reasoning, and color robustness. By contrast, our GenColorBench is tailored to evaluate color generation capabilities, a distinct focus that targets generative models and unified models. Notably, recent popular generative and unified models often build on existing LLMs or VLMs; this innovates an interesting future direction: considering both benchmarks to explore whether improvements in color generation tasks can enhance performance on deterministic color understanding tasks.

\section{Color Evaluation Framework}

\begin{figure*}[t]
    \centering
    \includegraphics[width=\linewidth]{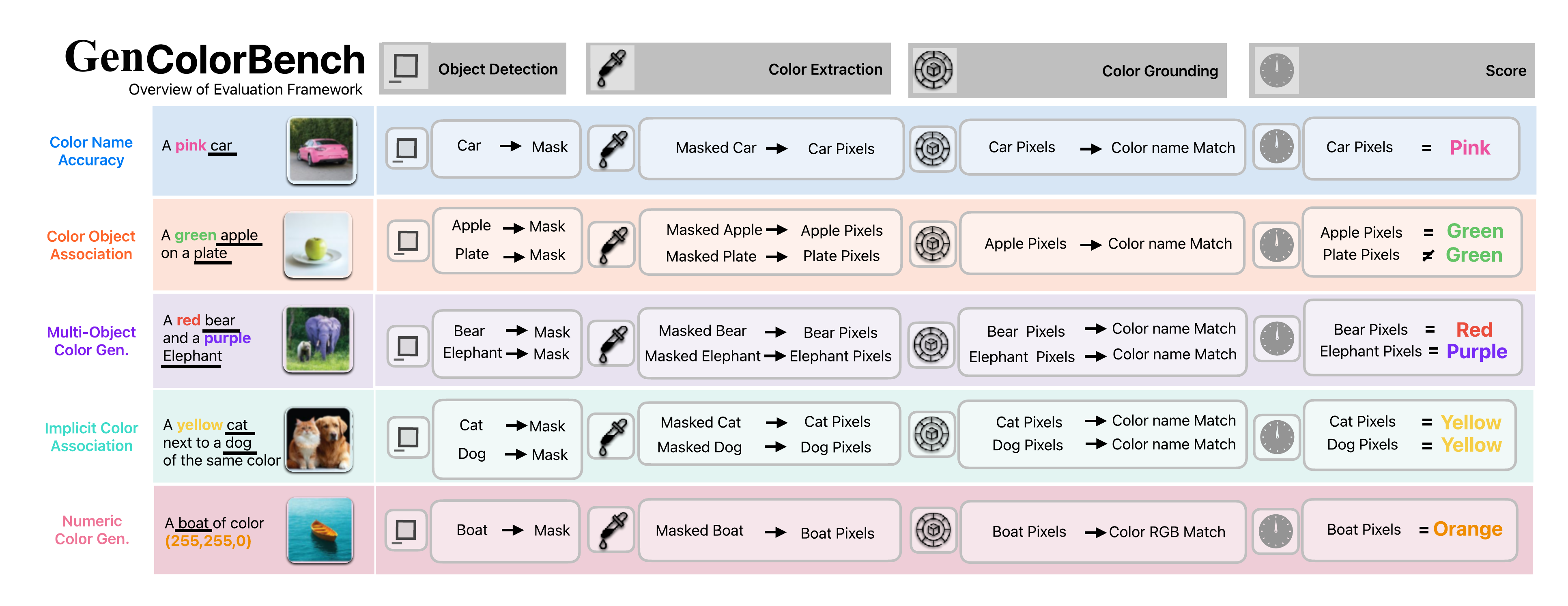}
    \vspace{-6mm}
    \caption{An overview of GenColorBench evaluation framework. The evaluation pipeline consists of five key components: VQA-based object localization, object segmentation, pixel extraction, color grounding, and score mechanism. Then, five color evaluation tasks are devised to analyse different aspects of color understanding in T2I models covering single object coloring, color-object association, multi-object color composition, numerical color understanding, and Implicit Color Association.}
    \vspace{-3mm}
    \label{fig:method}
    \vspace{-3mm}
\end{figure*}

\subsection{T2I Color Generation Tasks.}
Our primary goal is to evaluate unified vision-language and T2I models' ability to understand and generate images given explicit color prompts. We organize evaluation into multiple tasks, targeting different dimensions of color understanding, considering the practical use-cases for generative models. GenColorBench consists of five color evaluation tasks: (i) \textbf{Color Name Accuracy}---assesses whether the model correctly renders an object in the color specified by its linguistic name. (ii) \textbf{Color-Object Association}---evaluates whether the specified color is assigned to the correct object without erroneous attribution to contextual elements. (iii) \textbf{Multi-Object Color Composition}---assess correct color-object associations when multiple objects and corresponding color names are specified. (iv) \textbf{Implicit Color Association}---evaluates understanding of semantic relationships when a color is assigned to only one object but should also correspond to other objects. (v) \textbf{Numerical Color Understanding}---assesses comprehension of RGB triplets and hex codes for accurate color generation.

\vspace{-3mm}

\subsection{Color Taxonomy} 
Colors can be specified in text prompts in various ways---most commonly through linguistic color names such as "a red rose", but also through numerical codes such as hexadecimals (e.g., \texttt{\#ff0000}) or RGBs (e.g., (255, 0, 0)). These color expressions are often interpreted differently by the T2I models depending on their text encoders. Therefore, it is important to consider both the linguistic and numerical color representations to perform an in-depth evaluation of T2I models for color generation tasks. To this end, we ground our evaluation in two standard color naming systems i.e., ISCC-NBS, and CSS/X11 which offers human-understandable names along with their numerical representations. 

The ISCC-NBS~\citep{kelly1976color} is derived from the Munsell color system~\citep{munsell2022color} that is a perceptually uniform color space designed to align with the human color perception. Munsell's color system organizes colors along three perceptual axes, which are hue, value (lightness), and chroma (saturation), determined by empirical human experiments. ISCC-NBS discretizes this continuous color space into named categories, resulting in a three-level hierarchy of colors, ranging from coarse to fine-grained colors. Level 1 includes 13 broad color categories corresponding to basic color linguistic names such as green, red, or blue. Level 2 expands these 13 colors to 29 intermediate hues by incorporating modifiers such as light, deep, or strong. Level 3 provides fine-grained color names with precise distinctions, such as light bluish green or moderate purplish pink. We also use CSS3/X11 color set~\citep{w3c_csscolor}, which includes 147 colors that are widely used in web design and digital interfaces. These color names precisely map to both RGB and hexadecimal color values, making them ideal to be used in text-prompts for T2I color generation evaluation tasks. 

\subsection{Data Curation}
After establishing the color evaluation tasks and the color sets, we generate prompts for each color evaluation task. The data curation involves four key components: object selection, prompt template creation and categorization, integration of standardized colors, and human-in-the-loop quality assessment. Each component is designed to ensure that the generated prompts and the associated evaluation settings are grounded, scalable, and suitable for automated and human evaluation.

\minisection{Object Selection.}
We curate a set of 108 objects that span multiple semantic categories to ensure comprehensive coverage of color-object combinations.
These objects are drawn from two widely used datasets---COCO~\citep{lin2014microsoft}, and ImageNet~\citep{deng2009imagenet}, and grouped them into seven semantic domain including fruits and vegetables, tools and miscellaneous items, vehicles, animals, clothing and accessories, furniture and household objects, and sports and toys. Each object is selected based on recognizability in T2I generation, color variability for plausible appearance, and suitability for the segmentation which is a crucial step in the downstream mask-based evaluation.

\minisection{Prompt Creation and Categorization.} We begin by pairing the objects and the color sets, resulting in a large pool of valid object-color combinations that serves as a seed inputs for the prompt generation. For each color-object pair, we use a pool of hand-crafted and GPT-4o generated prompt templates to produce the prompts, which are aligned with one of the four difficulty levels---shown in Table~\ref{tab2:prompt_cat}. Level 1 templates produce simple object focused prompts that describe a single colored object. These prompts are designed to evaluate the color name accuracy and numerical color understanding task.
Level 2 templates embed the object within a contextual scene which are used for color name accuracy and color-object association task. Level 3 templates describe the scene involving more than two objects along with their corresponding colors to assess the multi-object color compositions. Level 4 templates describe semantically complex scenes having one object with the assigned color, while a second object is referring to the color of the first object.

\minisection{Quality Assessment.}
After completing prompt generation, we perform human-in-the-loop validation to ensure the linguistic quality and semantic clarity of the generated prompts. The prompts are reviewed for grammatical check, 
 and ambiguity, especially in scene descriptive and implicit color association prompts. A random subset of prompts from each set are picked for review to ensure that the color references are unambiguous and the prompt structure does not mislead the models. All the ambiguous prompts are either revised or removed from the final sets.

\minisection{Prompt Distribution.}
Finally, we get 18K object focused prompts with linguistic color names, and 11.5K prompts with numerical colors including hex codes and RGB triples. The contextual object category includes 8.7K prompts to assess the object-color association. To evaluate multiple object generation, the scene descriptive category contains 2.2K prompts that embed colors within broader contexts. The implicit color association category includes 4.5K prompts where color attributes must be inferred based on semantic relationships between objects. This prompt distribution ensures a comprehensive evaluation of color grounding across a wide range of complexity levels, resulting into a large-scale set of 44K+ prompts. To facilitate broader accessibility and reproducibility, we further curate a compact, representative subset of less than 10K prompts—carefully selected to preserve semantic diversity and evaluation fidelity---making it readily usable by the research community.

\vspace{-3mm}
\begin{table*}[t]
\centering
\begin{minipage}[t]{0.54\textwidth}
\centering
\resizebox{\linewidth}{!}{
\begin{tabular}{lcccccc}
\hline
 & \multicolumn{2}{c}{Open-Ended} & \multicolumn{2}{c}{MCQ} & \multicolumn{2}{c}{Binary} \\
\cline{2-7}
Model               & CSS & L2 & CSS & L2 & CSS & L2 \\
\hline
Janus 1.3B        & 5.03    & 25.86   & 12.20   & 33.99  & 30.42   & 34.98 \\
Janus-Pro 7B      & 6.62    & 26.60   & 19.44   & 43.60  & 24.98   & 37.19 \\
mPLUG-Owl3 7B     & 7.24    & 24.14   & 17.93   & 42.12  & 26.87   & 41.87 \\
DeepSeek-VL2-7B   & 11.35   & 27.34   & 18.85   & 45.32  & 31.24   & 42.12 \\
BLIP3o-8B         & 12.17   & 25.12   & 24.73   & 45.81  & 31.10   & 44.09 \\
Qwen2-VL-7B       & 9.35    & 24.63   & 23.23   & 43.35  & 35.13   & 49.01 \\
Instruct-VL-7B    & 7.19    & 26.60   & 20.55   & 45.57  & 31.15   & 41.63 \\
\hline
Ours           & \multicolumn{3}{c|}{\textbf{L2: 96.46}} & \multicolumn{3}{c}{\textbf{CSS: 92.00}} \\
\hline
\end{tabular}
}
\caption{Performance (accuracy) of VLMs-based VQA on CSS/X11 and ISCC-NBS Level 2 colors.}
\label{tab:vqa_results}
\end{minipage}%
\hfill
\begin{minipage}[t]{0.44\textwidth}
\centering
\resizebox{\linewidth}{!}{
\begin{tabular}{p{1.5cm}|c|p{4.3cm}}
\hline
Type & \# Temp. & Example Prompt \\
\hline
Object-Focused & 12 & \textit{a red apple} \\
\hline
Contextual Object & 62 & \textit{a red apple on a white plate} \\
\hline
Scene Descriptive & 30 & \textit{a red apple on a white plate placed on a kitchen shelf} \\
\hline
Implicit Color Association & 100 & \textit{a red apple on a plate placed on a kitchen shelf. The plate is of the same color as the apple.} \\
\hline
\end{tabular}
}
\caption{Prompt categorization across four levels of difficulty, from simple to complex.}
\label{tab2:prompt_cat}
\end{minipage}
\vspace{-6mm}
\end{table*}

\subsection{Evaluation Framework}

\minisection{Object Detection.}
Our framework comprises three key components: object detection and segmentation, color grounding, and scoring mechanism to ensure object-aware perceptually aligned assessment. Following the Davidsonian Scene Graph (DSG) framework~\citep{cho2023davidsonian}, we employ Visual Question Answering (VQA)-based validation to first confirm the presence of the intended object(s) in the generated image before proceeding to attribute-level assessments such as color. For instance, given an input image along with ground truth, we formulate binary queries such as "Is there a car in the image", and rely on VQA response to determine the existence of object. For the multi-object tasks, the VQA model is queried for each object separately, and the image is validated only if all the objects in text prompts are present in the image. This ensures object-level precision in the evaluation tasks, especially in those that involve color association and color grounding between multiple objects. In practice, after empirical testing across several VLMs, we employ Janus-1.3B as VQA model due to its favorable trade-off between computational efficiency and reliability.

Then, a binary mask of the object is generated for color extraction. We use Grounded SAM ~\citep{ren2024grounded} pipeline which uses grounding DINO for text guided coarse localization of object and then SAM is used to produce final mask. Another reason for employing Grounded SAM is that the object may contain additional associated regions not required for the color grounding i.e., a mask of car may include lights, and wind shields that are not required in the color grounding. We refer these components as negative labels, and generated a list of the negative labels for all the objects using GPT-4o. To remove these negative objects from the mask, we apply negative Intersection-over-Union (IoU) filtering over positive mask to ensure separation of spatial region of the object. 

\minisection{Color Grounding and Score Mechanism.}
We propose to use a perceptually grounded, multi-metric evaluation protocol. 
Instead of direct color metrics like DeltaE that penalize lighting variations, we extract RGB pixels from predicted masks and transform them to CIELAB space  denoted as $\mathbf{P}$ = $\left( L_i^*, a_i^*, b_i^* \right)_{i=1}^N$. The object may exhibit polychromatic color distribution due to geometric and lighting variations, but human observers typically abstract these variations, attributing a single representative color to an object. To capture this fundamental aspect of human vision, we adopt the dominant hue concept which is explored by~\citep{WITZEL2022}, which identifies the representative color of an object by focusing on primary direction of chromatic variation within its color distribution.
Then, we perform principal component analysis on the chromatic components ($a*$ and $b*$) of the CIELAB pixel values. It is noted by~\citep{WITZEL2022} that the first component $\mathbf{v}_1 = (v_{1a}, v_{1b})$ of chromaticity distribution $\mathbf{P}_{ab} = \left( a_i^*, b_i^* \right)_{i=1}^N$ represents the dominant hue. Then, chromaticity of $a_i^*, b_i^*$ is projected onto this dominant hue direction $\mathbf{v}_1$ and mean of lightness ($\overline{L}^*$) and the projected chromatic values ($\overline{a_{\text{proj}}^*}$, $\overline{b_{\text{proj}}^*}$) are computed to obtain the dominant color.

Now, we have the dominant color of the object and ground truth color from ISCC-NBS or CSS3/X11 color sets. However, a key challenge arises: can a single nominal color label—such as “pink” from ISCC–NBS Level 1—adequately represent the full perceptual gamut of that color category? In practice, a dominant color may correspond to a slightly different but perceptually indistinguishable shade. To account for this variability and avoid penalizing perceptually plausible matches, we construct a candidate set for each ground-truth color by including the nominal color along with its \textbf{\textit{k}} perceptually nearest neighbors in the same color-naming system.

We compute three complementary metrics: (i) Delta Chroma ---  the Euclidean distance in $a^*, b^*$ chromaticity plane, (ii) CIEDE2000 --- distribution level distance between in $L^*$, $a^*$, $b^*$ space, and (iii) MAE (Hue) --- an angular difference in hue, computed in polar coordinates with chroma-based reliability gating. For each metric, we compute the minimum perceptual distance between the predicted dominant color and the candidate set. This distance is compared against the metric-specific JND threshold (typically 5), with binary scores assigned based on whether the distance falls below the threshold. An overall "Correct" assessment requires all metrics to pass. 


\begin{table*}[t]
\centering
\resizebox{\linewidth}{!}{
\begin{tabular}{l|c|c|ccccc|c}

Model & Resolution & Type & Color Name & Color-Object & Multi-Object Color & Implicit Color  & Numerical Color & Avg.\\
 &  &                                  & Accuracy & Association & Composition & Association & Understanding & \\
\hline
Flux & 1024 & DM & 33.70 & 18.99 & \cellcolor[HTML]{fdd49f} 10.49 & \cellcolor[HTML]{fc8d58} 22.49 & 9.14 & 18.96\\
\hline
Sana & 1024 & DM & \cellcolor[HTML]{fc8d58} 49.85 & 18.10 & 7.06 & 15.18  & \cellcolor[HTML]{fff6ee} 15.80 & 21.20\\
\hline
SD 3.5 & 1024 & DM & \cellcolor[HTML]{fdd49f} 49.83 & \cellcolor[HTML]{fff6ee} 20.53 & \cellcolor[HTML]{fc8d58} 11.43 & 17.81 & 9.41 & \cellcolor[HTML]{fff6ee} 21.80\\
\hline
Pixart Alpha & 1024 & DM & \cellcolor[HTML]{fff6ee} 49.61 & 13.48 & 1.73 & 9.47 & 6.36 & 16.13 \\
\hline
SD 3 & 1024 & DM & 45.97 & \cellcolor[HTML]{fdd49f} 22.45 & 9.84 & 13.17  & 7.45 & 19.78 \\
\hline
Pixart Sigma & 1024 & DM & 47.36 & 16.75 & 3.05 &  11.49 & 6.47 & 17.02\\
\hline
Janus Pro & 384 & AR & 29.55 & 16.33 & 8.25 &  17.88  & 3.66 & 15.13\\
\hline
OmniGen2 & 512 & AR & 42.47 & \cellcolor[HTML]{fc8d58} 23.71 & \cellcolor[HTML]{fff6ee} 9.91 & \cellcolor[HTML]{fff6ee} 18.51 & \cellcolor[HTML]{fdd49f} 17.49 & \cellcolor[HTML]{fc8d58} 22.42 \\
\hline
Blip3o & 1024 & MM & 40.59 & 15.59 & 5.21 & \cellcolor[HTML]{fdd49f} 21.35 & \cellcolor[HTML]{fc8d58} 28.31 & \cellcolor[HTML]{fdd49f} 22.21\\

\end{tabular}
}

\caption{Overall performance of T2I models on GenColorBench. \textit{The scores are averaged over ISCC-NBS L2, L3, and CSS3/X11 colors.} \colorsquare{first} \colorsquare{second} \colorsquare{third} incidate best, second-best, and third-best.
}

\label{tab3:results}
\end{table*}
\vspace{-3mm}
\section{Benchmark}
Most existing benchmarks assess color fidelity in text-to-image generation using VQA-based approaches, as summarized in Table~\ref{tab1:existing_benchmarks}. However, these methods often rely on VLLMs that lack direct grounding in pixel-level color information, making them susceptible to hallucination, linguistic bias, and imprecise color perception. To rigorously evaluate this limitation, we constructed a controlled diagnostic set of 2464 synthetic images rendered in Blender using CSS3/X11 and ISCC–NBS L2 colors. We evaluated seven state-of-the-art VLLMs on three tasks: (i) open-ended color name/hex code prediction, (ii) multiple-choice RGB selection, and (iii) binary color verification.

As shown in Table~\ref{tab:vqa_results}, the best-performing VLLM (Qwen2-VL) achieves only 49.01\% accuracy on L2 binary task and 24.73\% on CSS MCQ task, with open-ended performance remaining critically low (below 12.17\%). These results confirm that current VLLMs struggle to reliably distinguish fine-grained colors, even under ideal conditions with single-object scenes. In contrast, our proposed method achieves 96.46\% accuracy on L2 and 92.00\% on CSS3/X11 colors (see appendix for details). 



\subsection{Experiment Setup}
\minisection{Models.}
We focus on a broad range of the recent T2I models. This includes Flux.1~\citep{flux2024}; Stable Diffusion 3.5~\citep{stabilityai_sd3_5_large_2024} and Stable Diffusion 3~\citep{stabilityai_sd3_medium_diffusers_2025} from the stability AI;  PixArt-$\alpha$~\citep{chen2023pixart} and PixArt-$\sigma$~\citep{chen2024pixart} from the PixArt family; autoregressive models such as Janus Pro~\citep{wu2025janus} and OmniGen2~\citep{wu2025omnigen2}; multimodal model BLIP3o~\citep{chen2025blip3}; and Sana~\citep{xie2024sana}---an optimized model for semantic and visual grounding. These models represent diverse architectures, ranging from diffusion-based pipelines to autoregressive and hybrid approaches. Further details are provided in the Appendix.

\begin{figure*}[t]
\centering
\includegraphics[width=\linewidth]{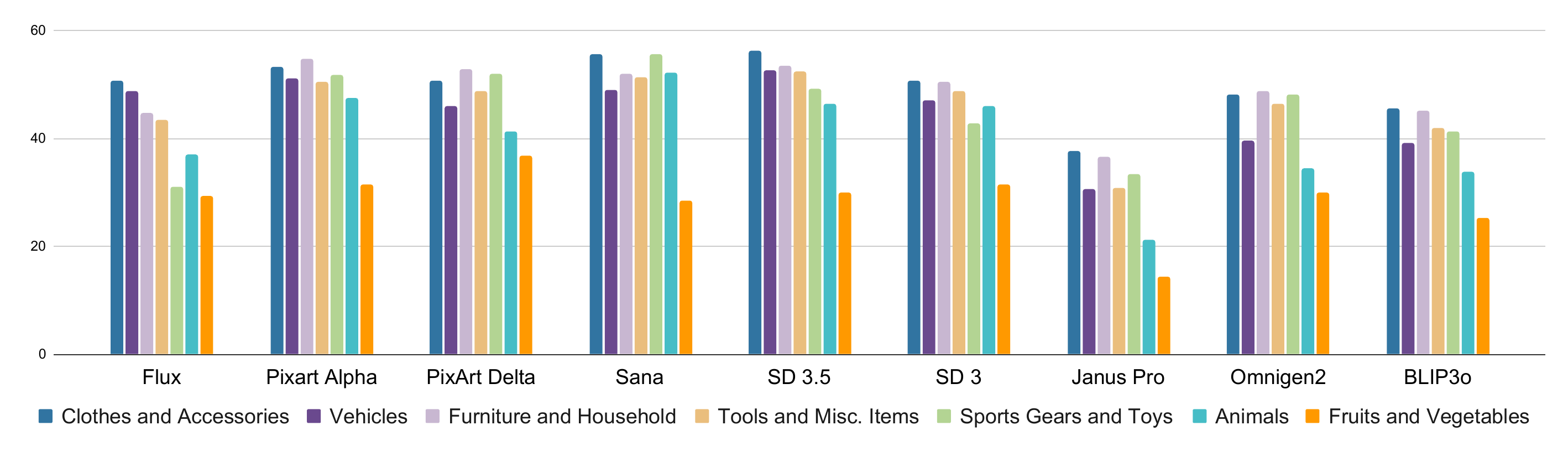}
\vspace{-7mm}
\caption{Performance of T2I model on category-wise color accuracy. \textit{The scores are averaged over the Level 2 and Level 3 ISCC-NBS colors, and CSS3/X11 colors based object focused prompts.}}
\label{fig:cat_accuracy}
\vspace{-4mm}
\end{figure*}

\minisection{Image Generation.}
The evaluation is performed on a set of 44,464 prompts spanning all the five tasks described in Table~\ref{tab2:prompt_cat}. Following the practice in existing benchmarks, we generate 4 images per prompt, and compute the average score across all the generated images. For each model, the hyper-parameters including sampling step, and image resolution are set to default to ensure fairness in comparison. Image generation is performed using Nvidia A40 GPUs.

\subsection{Overall Performance} We evaluate the performance of various T2I models on five color generation tasks using GenColorBench, with results summarized in Table~\ref{tab3:results}. For each task, scores are averaged across color prompts derived from Levels 2 and 3 of the ISCC-NBS system and CSS/X11 color names. Despite architectural diversity — including diffusion models (DM), autoregressive models (AR), and multimodal architectures (MM) — all models exhibit a consistent trend: performance degrades as task complexity increases. OmniGen2~\citep{wu2025omnigen2} achieves the highest average score (22.42), followed closely by BLIP3o (22.21) and Stable Diffusion 3.5 (21.80). Notably, OmniGen2 operates at a lower resolution (512×512) compared to SD 3.5 and BLIP3o (both 1024×1024), suggesting its superior performance is not merely resolution-dependent but may reflect stronger color semantics modeling.

\begin{figure*}[t]
    \centering
    \includegraphics[width=0.97\linewidth]{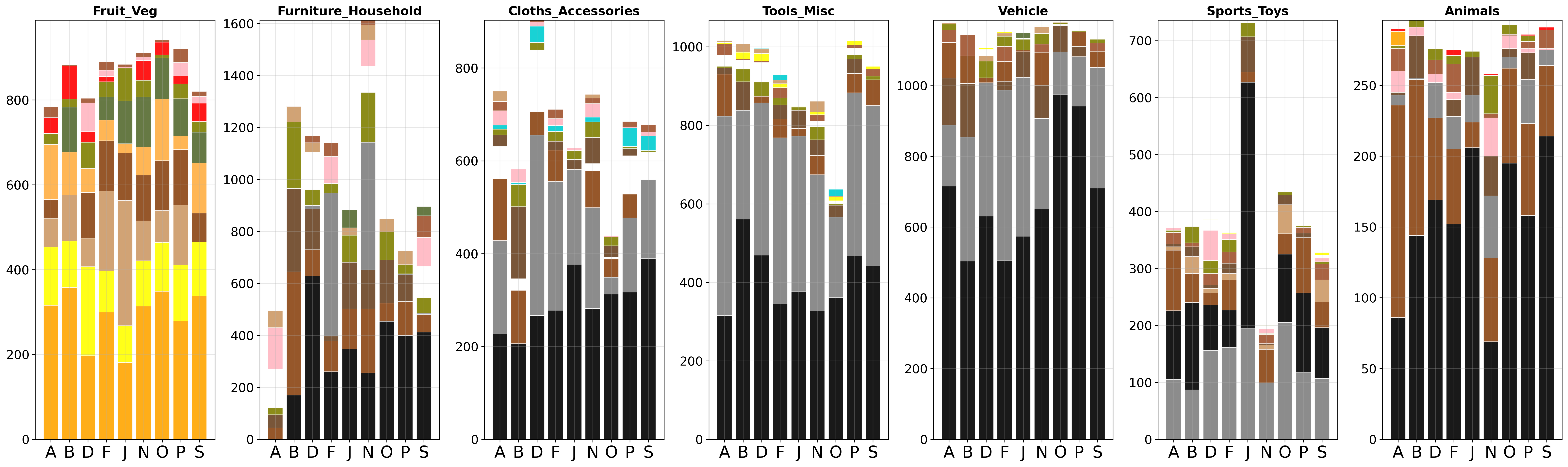}
    \caption{Distribution of estimated dominant colors (Top-10) across 10,000 generated images for each T2I models, revealing inherent color biases in vanilla baseline models. Models include: A = PixArt Alpha, B = BLIP3o, F = Flux, J = Janus-Pro,  N = Sana, O = OmniGen2,  P = PixArt Sigma,  S = Stable Diffusion 3, and D = Stable Diffusion 3.5. \textit{Interestingly, all the models are significantly biased towards black, gray, and brown across all the categories except fruits and vegetables.}}
    \label{fig:bias_analysis}
\end{figure*}

\begin{figure*}[t]
    \centering
    \includegraphics[width=0.97\linewidth]{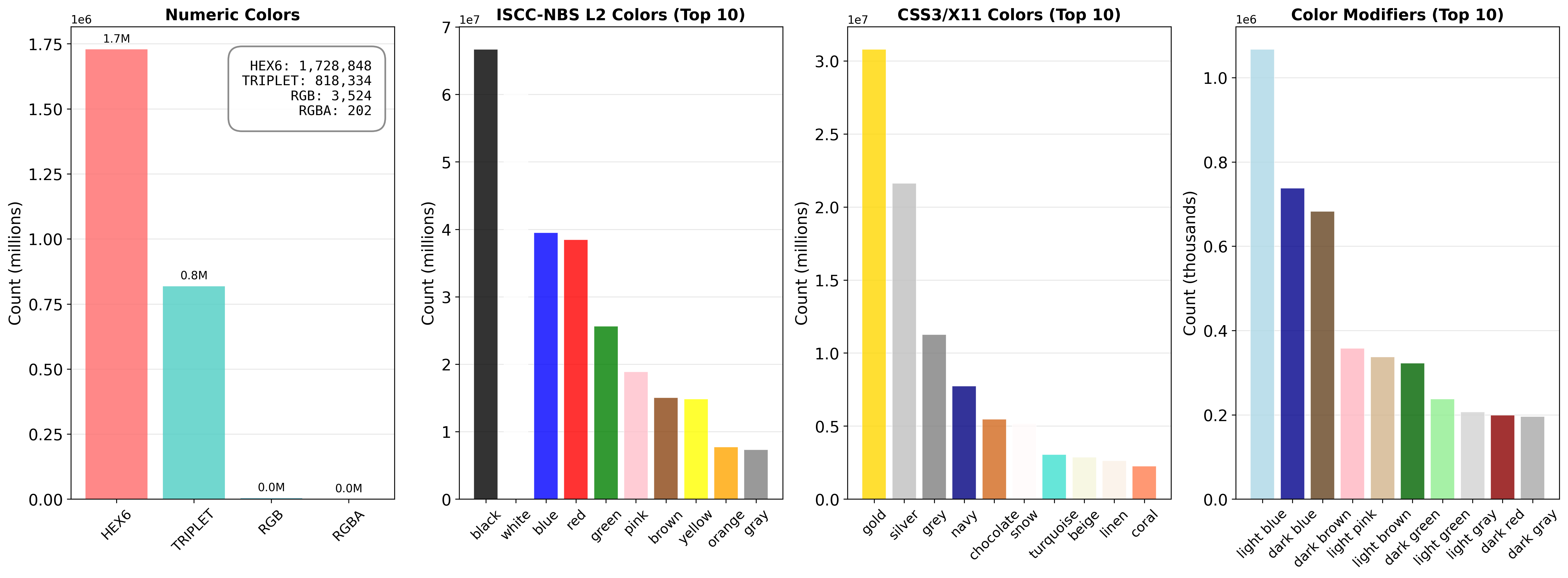}
    \vspace{-3mm}
    \caption{Color representation in LAION-2B text prompts, analyzed across four semantic categories: (i) Numeric Colors, (ii) ISCC-NBS L2 colors, (iii) CSS3/X11 named colors, and (iv) Color Modifiers. \textit{The data reveals the dominant representation of ISCC-NBS L2 colors and their modifiers. Whereas, the numeric colors are significantly under-represented as compared to the named colors.}}
    \label{fig:laion_analysis}
    \vspace{-5mm}
\end{figure*}

On task-specific metrics, Stable Diffusion 3.5 (49.83) and Sana (49.85) lead in Color Name Accuracy, indicating strong grounding of color names, though even top performers remain below 50\%, revealing persistent difficulty with fine-grained or ambiguous color terms. In contrast, performance plummets in the Color-Object Association task, where only OmniGen2 exceeds 23\% (23.71), underscoring widespread failure in assigning colors to specific objects without leakage or misattribution. The Multi-Object Color Composition task reveals a sharp drop in performance across all models — with scores generally below 12 — highlighting severe limitations in spatially disentangling and assigning distinct colors to multiple objects simultaneously. Similarly, in the Implicit Color Association task, models struggle to infer color relationships embedded in texture, context, or scene semantics, with scores rarely exceeding 23\%. Finally, the Numerical Color Understanding task proves most challenging, with most models scoring under 10\%. Interestingly, BLIP3o significantly outperforms others here (28.31), suggesting its multimodal architecture may better encode or reason about explicit numeric color representations (e.g., RGB/hex values), which are typically learned implicitly in conventional T2I pipelines. These results collectively demonstrate that while modern T2I models can approximate basic color naming, they remain fundamentally limited in their ability to precisely control, associate, or numerically interpret color within complex visual compositions.

\vspace{-3mm}
\subsection{Category-Level Analysis}
We evaluate how T2I models ground color names across seven semantic object categories as shown in Figure~\ref{fig:cat_accuracy}. A clear pattern emerges: models consistently achieve higher accuracy on categories such as \textit{Clothes and Accessories}, \textit{Vehicles}, and \textit{Furniture and Household}, where color is often stylistic or decorative rather than semantically bound to identity. In contrast, performance drops sharply for \textit{Animals} and \textit{Fruits and Vegetables}, where color is biologically intrinsic (e.g., yellow banana) and requires precise disentanglement of object identity from color attribute. This disparity reflects a deep-seated training data biases. As revealed in Figure~\ref{fig:bias_analysis}, all models exhibit strong chromatic bias toward \textit{black}, \textit{gray}, and \textit{brown} across nearly all categories, mirroring the dominant color distribution observed in LAION-2B text prompts in Figure~\ref{fig:laion_analysis}. Notably, neutral tones are overrepresented in training corpora, particularly in Vehicles and Furniture category, which explains models’ relative success there. Conversely, vibrant or biologically specific colors such as reds, yellows are underrepresented in both training prompts and generated outputs, especially for Animals, and Fruits and Vegetables.

This alignment between model output bias and dataset statistics suggests that current T2I systems largely rely on statistical co-occurrence patterns rather than compositional reasoning about color semantics. For instance, the persistent rendering of bananas as “yellow” stems not from learning biological color norms, but from memorizing frequent associations in the training corpus — a phenomenon consistent with prior findings on human color-concept associations~\citep{rathore2019estimating}.
OmniGen2 and Stable Diffusion 3.5 show better cross-category generalization, while Janus Pro and BLIP3o exhibit the weakest performance, particularly struggling with color control in biologically constrained categories. This highlights that compositional color control remains challenging when decoupling color from object identity.

\vspace{-3mm}

\subsection{Basic and Intermediate Color Understanding}
We evaluate T2I models on basic and intermediate color understanding. To achieve this, we categorize the Red, Orange, Brown, Yellow, Olive, Yellow, Green, Blue, Purple, White, Gray, and Black as basic colors ---similar to conventional color naming approaches \citep{berlin1991basic} where colors are described with a single word. We then group all the rest of Level 2 colors as intermediate colors. 
We measure the accuracy of these categories using the color naming accuracy task and illustrate the results in Figure~\ref{fig:color_comparisons}(Left). These results indicate that all models perform well on basic colors, but consistently struggle with intermediate color grounding, which proves to be a more difficult task. Interestingly, there is not a large difference in the order of the models with both sets of colors, being Sana, Stable Diffusion 3.5, and PixArt-Alpha the ones obtaining best results for both type of colors.



\vspace{-3mm}
\subsection{Modifier-based Compositionality}
\begin{figure}[t]
\centering
\begin{subfigure}[t]{0.47\linewidth}
    \centering
    \includegraphics[width=\linewidth]{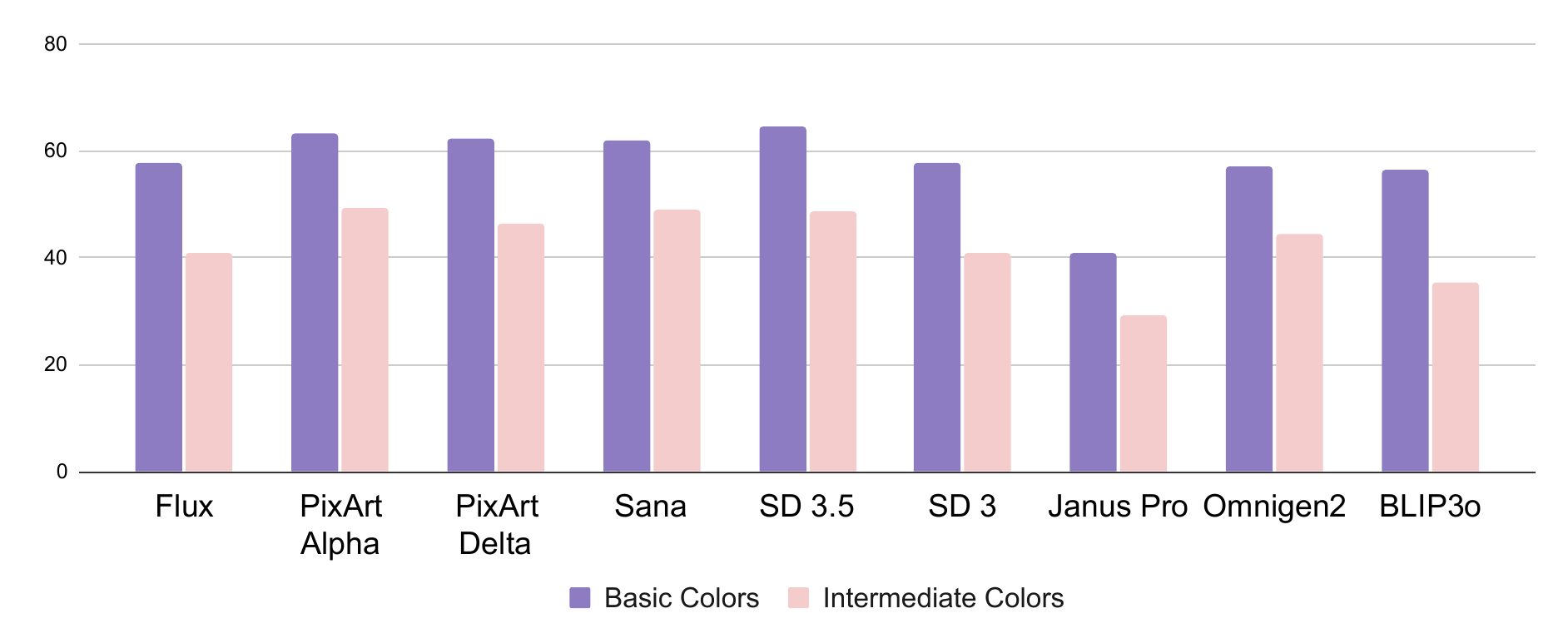}
    \label{fig:solid_inter_col}
\end{subfigure}%
\hfill
\begin{subfigure}[t]{0.47\linewidth}
    \centering
    \includegraphics[width=\linewidth]{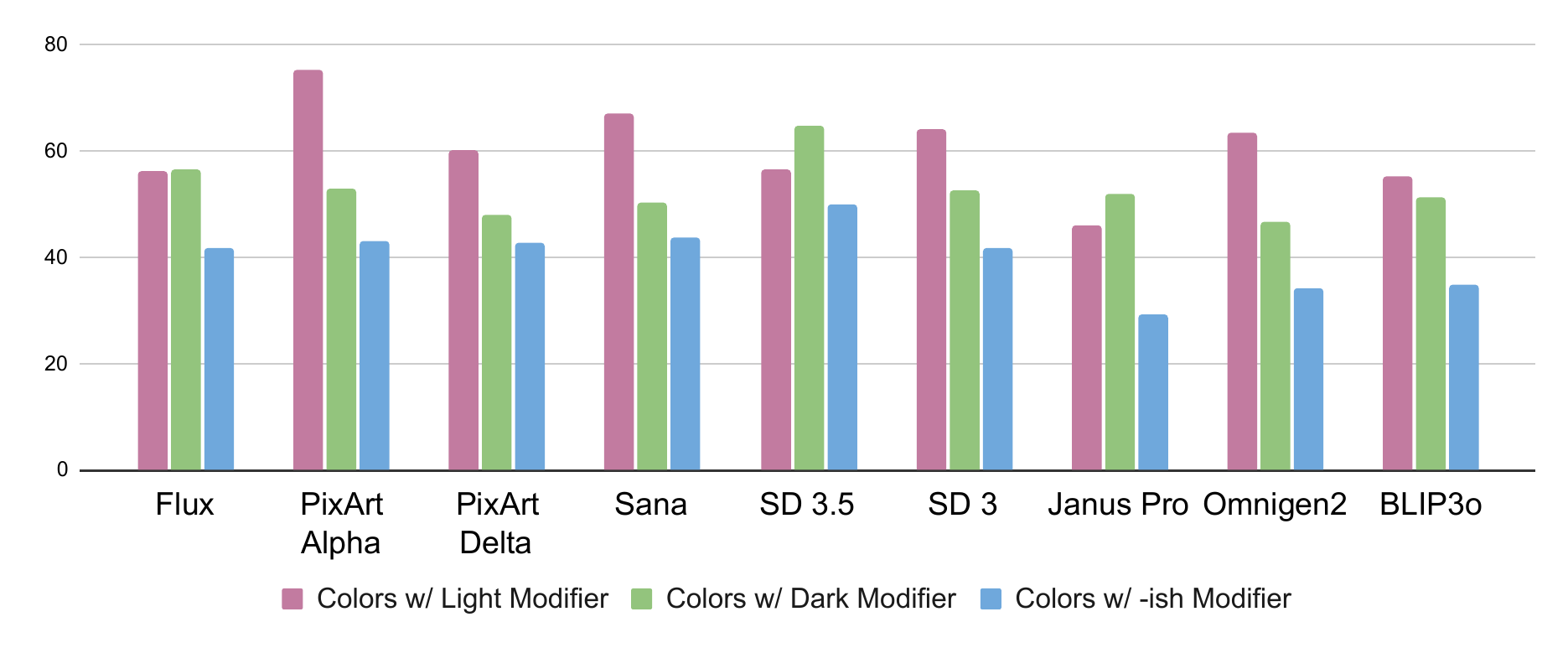}
    \label{fig:modifier_col}
\end{subfigure}
\vspace{-6mm}
\caption{(Left) Comparison b/w basic and intermediate colors. These models better understand basic colors, while accuracy drops by 8--20\% on intermediate colors. (Right) Comparison of color modifiers. These models understand light color modifiers better, while -ish modifiers remain worst.}

\label{fig:color_comparisons}
\vspace{-4mm}
\end{figure}

We also analyse the understanding of color modifiers (i.e., dark, light, -ish) in T2I models. These modifiers are commonly used in natural languages to define different variants of the basic colors, e.g. light blue, dark blue, and greenish blue. Therefore, we group the ISCC-NBS Level 3 colors based on these three modifiers and study the color name accuracy task for each group. The results in terms of accuracy are shown in Figure~\ref{fig:color_comparisons}(Right) which demonstrate that these models perform better with light modified colors, as compared to the dark modified colors. On the other hand, -ish modified colors remain a hard task for all the models with the performance often below than 35\%, highlighting that these models struggle with gradient color semantics described in natural language. 



\vspace{-3mm}
\section{Conclusions}

We introduce GenColorBench, the first comprehensive benchmark for assessing color generation accuracy of T2I models. Our analysis of state-of-the-art models and reveals significant limitations in their ability to adhere to precise color specifications, highlighting the need for improved color controllability. GenColorBench’s focus on both categorical color names and numerical values (RGB, hex) fills a key void in existing evaluation frameworks, providing a robust tool for measuring progress in this essential dimension. By establishing baseline metrics and identifying failure modes, this work lays groundwork for advancing T2I models’ fidelity to color prompts. 

\section*{Acknowledgements}

This work was supported by Grants PID2021-128178OB-I00, PID2022-143257NB-I00, and PID2024-162555OB-I00 funded by MCIN/AEI/10.13039/ 501100011033 and FEDER, by the Generalitat de Catalunya CERCA Program, by the grant Càtedra ENIA UAB-Cruïlla (TSI-100929-2023-
2) from the Ministry of Economic Affairs and Digital
Transition of Spain, and the ELLIOT project Funded by the European Union ELLIOT project.
JVC also acknowledges the 2025 Leonardo Grant for Scientific Research and Cultural Creation from the BBVA Foundation. The BBVA Foundation accepts no responsibility for the opinions, statements and contents included in the project and/or the results thereof, which are entirely the responsibility of the authors.
Kai Wang acknowledges the funding from Guangdong and Hong Kong Universities 1+1+1 Joint Research Collaboration Scheme and the start-up grant B01040000108 from CityU-DG.

\bibliography{iclr2026_conference}
\bibliographystyle{iclr2026_conference}

\clearpage

\appendix

\section{Appendix: Statements}

\paragraph{Limitations/future work.}
Our benchmark is grounded in English-based color naming systems (ISCC-NBS and CSS/X11), which may not fully capture cross-linguistic variations in color conceptualization~\citep{lindner2012large},
The choice of English is motivated by its role as a \textit{lingua franca} in both large-scale dataset curation and the development of foundational generative models, most of which are trained predominantly on English-aligned web data. Despite this, we believe GenColorBench provides a crucial first step toward comprehensive color evaluation frameworks and establishes essential baseline metrics that can guide future research in improving color controllability in generative models.

\paragraph{Broader Impacts.}
GenColorBench enhances the flexible stylization capability in text-to-image synthesis by disentangling the color and texture elements. However, it also carries potential negative implications. It could be used to generate false or misleading images, thereby spreading misinformation. If  is applied to generate images of public figures, it poses a risk of infringing on personal privacy. Additionally, the automatically generated images may also touch upon copyright and intellectual property issues.

\paragraph{Ethical Statement.}
We acknowledge the potential ethical implications of deploying generative models, including issues related to privacy, data misuse, and the propagation of biases. All models used in this paper are publicly available. We will release the modified codes to reproduce the results of this paper. 
We also want to point out the potential role of customization approaches in the generation of fake news, and we encourage and support responsible usage. 

\paragraph{Reproducibility Statement.}
To facilitate reproducibility, we will make the entire source code and scripts needed to replicate all results presented in this paper available after the peer review period. We will release the code for the novel color metric we have introduced. We conducted all experiments using publicly accessible datasets. Elaborate details of all experiments have been provided in the Appendices. 

\paragraph{LLM usage statement.} 
We used a large language model solely to aid in polishing the writing and improving the clarity of the manuscript. The model was not involved in ideation, data analysis, or deriving any of the scientific contributions presented in this work.

\section{Details of Color Taxonomy.}
To ensure a standardized evaluation of color fidelity in text-to-image generation, we ground our analysis in two widely recognized color naming systems: the Inter-Society Color Council – National Bureau of Standards (ISCC–NBS) system and the CSS3/X11 colors. These color systems are well-established and offer both perceptually meaningful color categories and precise numerical representations, facilitating human-aligned assessments.

The ISCC–NBS system organizes colors hierarchically, making it suitable for both coarse and fine-grained color evaluation. Specifically, we use Level 2 of the ISCC–NBS color system, which comprises 29 basic color categories. These represent broad, commonly recognized color terms grounded in perceptual uniformity. Moreover, we also use Level 3 colors, a finer-grained extension consisting of 267 distinct color names that serve as subcategories of the Level 2 set. This color set provides variation (e.g., moderate red, deep yellowish green, light bluish purple) and allow us to examine the generative models' sensitivity to subtle differences in hue, saturation, and brightness.

In addition, we incorporate the CSS3/X11 color specification, a standard widely used in web development. This set consists of 147 named colors (e.g., dodgerblue, crimson, darkslategray), each with predefined RGB values and hex color codes. These color names are familiar to a broad audience and offer an alternative taxonomy that complements the ISCC–NBS system with prevalent color terms.

We provide detailed summaries of all the color sets used in our evaluation. The ISCC–NBS Level 2 colors are listed in Table~\ref{tab:l2_colors}. ISCC–NBS Level 3 colors are listed in Table~\ref{tab:l3_colors}. While, the CSS3/X11 color set along with their RGB and hex codes are listed in Table~\ref{tab:css_colors}.

\begin{table}[!t]
\centering
\begin{tabular}{|c|l|c|c|c|c|}
\hline
\textbf{ID} & \textbf{Color Name} & \textbf{R} & \textbf{G} & \textbf{B} & \textbf{Color} \\
\hline
1  & Pink             & 230 & 134 & 151 & \cellcolor[RGB]{230,134,151} \\
2  & Red              & 185 & 40  & 66  & \cellcolor[RGB]{185,40,66}   \\
3  & Yellowish pink   & 234 & 154 & 144 & \cellcolor[RGB]{234,154,144} \\
4  & Reddish orange   & 215 & 71  & 42  & \cellcolor[RGB]{215,71,42}   \\
5  & Reddish brown    & 122 & 44  & 38  & \cellcolor[RGB]{122,44,38}   \\
6  & Orange           & 220 & 125 & 52  & \cellcolor[RGB]{220,125,52}  \\
7  & Brown            & 127 & 72  & 41  & \cellcolor[RGB]{127,72,41}   \\
8  & Orange yellow    & 227 & 160 & 69  & \cellcolor[RGB]{227,160,69}  \\
9  & Yellowish brown  & 151 & 107 & 57  & \cellcolor[RGB]{151,107,57}  \\
10 & Yellow           & 217 & 180 & 81  & \cellcolor[RGB]{217,180,81}  \\
11 & Olive brown      & 127 & 97  & 41  & \cellcolor[RGB]{127,97,41}   \\
12 & Greenish yellow  & 208 & 196 & 69  & \cellcolor[RGB]{208,196,69}  \\
13 & Olive            & 114 & 103 & 44  & \cellcolor[RGB]{114,103,44}  \\
14 & Yellow green     & 160 & 194 & 69  & \cellcolor[RGB]{160,194,69}  \\
15 & Olive green      & 62  & 80  & 31  & \cellcolor[RGB]{62,80,31}    \\
16 & Yellowish green  & 74  & 195 & 77  & \cellcolor[RGB]{74,195,77}   \\
17 & Green            & 79  & 191 & 154 & \cellcolor[RGB]{79,191,154}  \\
18 & Bluish green     & 67  & 189 & 184 & \cellcolor[RGB]{67,189,184}  \\
19 & Greenish blue    & 62  & 166 & 198 & \cellcolor[RGB]{62,166,198}  \\
20 & Blue             & 59  & 116 & 192 & \cellcolor[RGB]{59,116,192}  \\
21 & Purplish blue    & 79  & 71  & 198 & \cellcolor[RGB]{79,71,198}   \\
22 & Violet           & 120 & 66  & 197 & \cellcolor[RGB]{120,66,197}  \\
23 & Purple           & 172 & 74  & 195 & \cellcolor[RGB]{172,74,195}  \\
24 & Reddish purple   & 187 & 48  & 164 & \cellcolor[RGB]{187,48,164}  \\
25 & Purplish pink    & 229 & 137 & 191 & \cellcolor[RGB]{229,137,191} \\
26 & Purplish red     & 186 & 43  & 119 & \cellcolor[RGB]{186,43,119}  \\
27 & White            & 231 & 225 & 233 & \cellcolor[RGB]{231,225,233} \\
28 & Gray             & 147 & 142 & 147 & \cellcolor[RGB]{147,142,147} \\
29 & Black            & 43  & 41  & 43  & \cellcolor[RGB]{43,41,43}    \\
\hline
\end{tabular}
\caption{ISCC-NBS L2 Color Names and RGB Values}
\label{tab:l2_colors}
\end{table}

\begin{table*}[!htbp]
\centering
\resizebox{\textwidth}{!}{
\begin{tabular}{|c|l|c|c|c||c|l|c|c|c||c|l|c|c|c|}
\hline
\textbf{ID} & \textbf{Color Name} & \textbf{R} & \textbf{G} & \textbf{B} & \textbf{ID} & \textbf{Color Name} & \textbf{R} & \textbf{G} & \textbf{B} & \textbf{ID} & \textbf{Color Name} & \textbf{R} & \textbf{G} & \textbf{B} \\
\hline
1 & Vivid pink & 253 & 121 & 146 & 90 & Grayish yellow & 200 & 177 & 139 & 179 & Deep blue & 17 & 48 & 116 \\
2 & Strong pink & 244 & 143 & 160 & 91 & Dark grayish yellow & 169 & 144 & 102 & 180 & Very light blue & 153 & 198 & 249 \\
3 & Deep pink & 230 & 105 & 128 & 92 & Yellowish white & 238 & 223 & 218 & 181 & Light blue & 115 & 164 & 220 \\
4 & Light pink & 248 & 195 & 206 & 93 & Yellowish gray & 198 & 185 & 177 & 182 & Moderate blue & 52 & 104 & 158 \\
5 & Moderate pink & 226 & 163 & 174 & 94 & Light olive brown & 153 & 119 & 54 & 183 & Dark blue & 23 & 52 & 89 \\
6 & Dark pink & 197 & 128 & 138 & 95 & Moderate olive brown & 112 & 84 & 32 & 184 & Very pale blue & 194 & 210 & 236 \\
7 & Pale pink & 239 & 209 & 220 & 96 & Dark olive brown & 63 & 44 & 16 & 185 & Pale blue & 145 & 162 & 187 \\
8 & Grayish pink & 203 & 173 & 183 & 97 & Vivid greenish yellow & 235 & 221 & 33 & 186 & Grayish blue & 84 & 104 & 127 \\
9 & Pinkish white & 239 & 221 & 229 & 98 & Brilliant greenish yellow & 233 & 220 & 85 & 187 & Dark grayish blue & 50 & 63 & 78 \\
10 & Pinkish gray & 199 & 182 & 189 & 99 & Strong greenish yellow & 196 & 184 & 39 & 188 & Blackish blue & 30 & 37 & 49 \\
11 & Vivid red & 213 & 28 & 60 & 100 & Deep greenish yellow & 162 & 152 & 18 & 189 & Bluish white & 225 & 225 & 241 \\
12 & Strong red & 191 & 52 & 75 & 101 & Light greenish yellow & 233 & 221 & 138 & 190 & Light bluish gray & 183 & 184 & 198 \\
13 & Deep red & 135 & 18 & 45 & 102 & Moderate greenish yellow & 192 & 181 & 94 & 191 & Bluish gray & 131 & 135 & 147 \\
14 & Very deep red & 92 & 6 & 37 & 103 & Dark greenish yellow & 158 & 149 & 60 & 192 & Dark bluish gray & 80 & 84 & 95 \\
15 & Moderate red & 177 & 73 & 85 & 104 & Pale greenish yellow & 230 & 220 & 171 & 193 & Bluish black & 36 & 39 & 46 \\
16 & Dark red & 116 & 36 & 52 & 105 & Grayish greenish yellow & 190 & 181 & 132 & 194 & Vivid purplish blue & 68 & 54 & 209 \\
17 & Very dark red & 72 & 17 & 39 & 106 & Light olive & 139 & 125 & 46 & 195 & Brilliant purplish blue & 128 & 136 & 226 \\
18 & Light grayish red & 180 & 136 & 141 & 107 & Moderate olive & 100 & 89 & 26 & 196 & Strong purplish blue & 83 & 89 & 181 \\
19 & Grayish red & 152 & 93 & 98 & 108 & Dark olive & 53 & 46 & 10 & 197 & Deep purplish blue & 42 & 40 & 111 \\
20 & Dark grayish red & 83 & 56 & 62 & 109 & Light grayish olive & 142 & 133 & 111 & 198 & Very light purplish blue & 183 & 192 & 248 \\
21 & Blackish red & 51 & 33 & 39 & 110 & Grayish olive & 93 & 85 & 63 & 199 & Light purplish blue & 137 & 145 & 203 \\
22 & Reddish gray & 146 & 129 & 134 & 111 & Dark grayish olive & 53 & 48 & 28 & 200 & Moderate purplish blue & 77 & 78 & 135 \\
23 & Dark reddish gray & 93 & 78 & 83 & 112 & Light olive gray & 143 & 135 & 127 & 201 & Dark purplish blue & 34 & 34 & 72 \\
24 & Reddish black & 48 & 38 & 43 & 113 & Olive gray & 88 & 81 & 74 & 202 & Very pale purplish blue & 197 & 201 & 240 \\
25 & Vivid yellowish pink & 253 & 126 & 93 & 114 & Olive black & 35 & 33 & 28 & 203 & Pale purplish blue & 142 & 146 & 183 \\
26 & Strong yellowish pink & 245 & 144 & 128 & 115 & Vivid yellow green & 167 & 220 & 38 & 204 & Grayish purplish blue & 73 & 77 & 113 \\
27 & Deep yellowish pink & 239 & 99 & 102 & 116 & Brilliant yellow green & 195 & 223 & 105 & 205 & Vivid violet & 121 & 49 & 211 \\
28 & Light yellowish pink & 248 & 196 & 182 & 117 & Strong yellow green & 130 & 161 & 43 & 206 & Brilliant violet & 152 & 127 & 220 \\
29 & Moderate yellowish pink & 226 & 166 & 152 & 118 & Deep yellow green & 72 & 108 & 14 & 207 & Strong violet & 97 & 65 & 156 \\
30 & Dark yellowish pink & 201 & 128 & 126 & 119 & Light yellow green & 206 & 219 & 159 & 208 & Deep violet & 60 & 22 & 104 \\
31 & Pale yellowish pink & 241 & 211 & 209 & 120 & Moderate yellow green & 139 & 154 & 95 & 209 & Very light violet & 201 & 186 & 248 \\
32 & Grayish yellowish pink & 203 & 172 & 172 & 121 & Pale yellow green & 215 & 215 & 193 & 210 & Light violet & 155 & 140 & 202 \\
33 & Brownish pink & 203 & 175 & 167 & 122 & Grayish yellow green & 151 & 154 & 133 & 211 & Moderate violet & 92 & 73 & 133 \\
34 & Vivid reddish orange & 232 & 59 & 27 & 123 & Strong olive green & 44 & 85 & 6 & 212 & Dark violet & 52 & 37 & 77 \\
35 & Strong reddish orange & 219 & 93 & 59 & 125 & Moderate olive green & 73 & 91 & 34 & 213 & Very pale violet & 208 & 198 & 239 \\
36 & Deep reddish orange & 175 & 51 & 24 & 126 & Dark olive green & 32 & 52 & 11 & 214 & Pale violet & 154 & 144 & 181 \\
37 & Moderate reddish orange & 205 & 105 & 82 & 127 & Grayish olive green & 84 & 89 & 71 & 215 & Grayish violet & 88 & 78 & 114 \\
38 & Dark reddish orange & 162 & 64 & 43 & 128 & Dark grayish olive green & 47 & 51 & 38 & 216 & Vivid purple & 185 & 53 & 213 \\
39 & Grayish reddish orange & 185 & 117 & 101 & 129 & Vivid yellowish green & 63 & 215 & 64 & 217 & Brilliant purple & 206 & 140 & 227 \\
40 & Strong reddish brown & 139 & 28 & 14 & 130 & Brilliant yellowish green & 135 & 217 & 137 & 218 & Strong purple & 147 & 82 & 168 \\
41 & Deep reddish brown & 97 & 15 & 18 & 131 & Strong yellowish green & 57 & 150 & 74 & 219 & Deep purple & 101 & 34 & 119 \\
42 & Light reddish brown & 172 & 122 & 115 & 132 & Deep yellowish green & 23 & 106 & 30 & 220 & Very deep purple & 70 & 10 & 85 \\
43 & Moderate reddish brown & 125 & 66 & 59 & 133 & Very deep yellowish green & 5 & 66 & 8 & 221 & Very light purple & 228 & 185 & 243 \\
44 & Dark reddish brown & 70 & 29 & 30 & 134 & Very light yellowish green & 197 & 237 & 196 & 222 & Light purple & 188 & 147 & 204 \\
45 & Light grayish reddish brown & 158 & 127 & 122 & 135 & Light yellowish green & 156 & 198 & 156 & 223 & Moderate purple & 135 & 94 & 150 \\
46 & Grayish reddish brown & 108 & 77 & 75 & 136 & Moderate yellowish green & 102 & 144 & 105 & 224 & Dark purple & 86 & 55 & 98 \\
47 & Dark grayish reddish brown & 67 & 41 & 42 & 137 & Dark yellowish green & 47 & 93 & 58 & 225 & Very dark purple & 55 & 27 & 65 \\
48 & Vivid orange & 247 & 118 & 11 & 138 & Very dark yellowish green & 16 & 54 & 26 & 226 & Very pale purple & 224 & 203 & 235 \\
50 & Strong orange & 234 & 129 & 39 & 139 & Vivid green & 35 & 234 & 165 & 227 & Pale purple & 173 & 151 & 179 \\
51 & Deep orange & 194 & 96 & 18 & 140 & Brilliant green & 73 & 208 & 163 & 228 & Grayish purple & 123 & 102 & 126 \\
52 & Light orange & 251 & 175 & 130 & 141 & Strong green & 21 & 138 & 102 & 229 & Dark grayish purple & 81 & 63 & 81 \\
53 & Moderate orange & 222 & 141 & 92 & 143 & Very light green & 166 & 226 & 202 & 230 & Blackish purple & 47 & 34 & 49 \\
54 & Brownish orange & 178 & 102 & 51 & 144 & Light green & 111 & 172 & 149 & 231 & Purplish white & 235 & 223 & 239 \\
55 & Strong brown & 138 & 68 & 22 & 145 & Moderate green & 51 & 119 & 98 & 232 & Light purplish gray & 195 & 183 & 198 \\
56 & Deep brown & 87 & 26 & 7 & 146 & Dark green & 22 & 78 & 61 & 233 & Purplish gray & 143 & 132 & 144 \\
57 & Light brown & 173 & 124 & 99 & 147 & Very dark green & 12 & 46 & 36 & 234 & Dark purplish gray & 92 & 82 & 94 \\
58 & Moderate brown & 114 & 74 & 56 & 148 & Very pale green & 199 & 217 & 214 & 235 & Purplish black & 43 & 38 & 48 \\
59 & Dark brown & 68 & 33 & 18 & 149 & Pale green & 148 & 166 & 163 & 236 & Vivid reddish purple & 212 & 41 & 185 \\
60 & Light grayish brown & 153 & 127 & 117 & 150 & Grayish green & 97 & 113 & 110 & 237 & Strong reddish purple & 167 & 73 & 148 \\
61 & Grayish brown & 103 & 79 & 72 & 151 & Dark grayish green & 57 & 71 & 70 & 238 & Deep reddish purple & 118 & 26 & 106 \\
62 & Dark grayish brown & 62 & 44 & 40 & 152 & Blackish green & 31 & 42 & 42 & 239 & Very deep reddish purple & 79 & 9 & 74 \\
63 & Light brownish gray & 146 & 130 & 129 & 153 & Greenish white & 224 & 226 & 229 & 240 & Light reddish purple & 189 & 128 & 174 \\
64 & Brownish gray & 96 & 82 & 81 & 154 & Light greenish gray & 186 & 190 & 193 & 241 & Moderate reddish purple & 150 & 88 & 136 \\
65 & Brownish black & 43 & 33 & 30 & 155 & Greenish gray & 132 & 136 & 136 & 242 & Dark reddish purple & 95 & 52 & 88 \\
67 & Brilliant orange yellow & 255 & 190 & 80 & 156 & Dark greenish gray & 84 & 88 & 88 & 243 & Very dark reddish purple & 63 & 24 & 60 \\
68 & Strong orange yellow & 240 & 161 & 33 & 157 & Greenish black & 33 & 38 & 38 & 244 & Pale reddish purple & 173 & 137 & 165 \\
69 & Deep orange yellow & 208 & 133 & 17 & 158 & Vivid bluish green & 19 & 252 & 213 & 245 & Grayish reddish purple & 134 & 98 & 126 \\
70 & Light orange yellow & 252 & 194 & 124 & 159 & Brilliant bluish green & 53 & 215 & 206 & 246 & Brilliant purplish pink & 252 & 161 & 231 \\
71 & Moderate orange yellow & 231 & 167 & 93 & 160 & Strong bluish green & 13 & 143 & 130 & 247 & Strong purplish pink & 244 & 131 & 205 \\
72 & Dark orange yellow & 195 & 134 & 57 & 162 & Very light bluish green & 152 & 225 & 224 & 248 & Deep purplish pink & 223 & 106 & 172 \\
73 & Pale orange yellow & 238 & 198 & 166 & 163 & Light bluish green & 95 & 171 & 171 & 249 & Light purplish pink & 245 & 178 & 219 \\
74 & Strong yellowish brown & 158 & 103 & 29 & 164 & Moderate bluish green & 41 & 122 & 123 & 250 & Moderate purplish pink & 222 & 152 & 191 \\
75 & Deep yellowish brown & 103 & 63 & 11 & 165 & Dark bluish green & 21 & 75 & 77 & 251 & Dark purplish pink & 198 & 125 & 157 \\
76 & Light yellowish brown & 196 & 154 & 116 & 166 & Very dark bluish green & 10 & 45 & 46 & 252 & Pale purplish pink & 235 & 200 & 223 \\
77 & Moderate yellowish brown & 136 & 102 & 72 & 168 & Brilliant greenish blue & 45 & 188 & 226 & 253 & Grayish purplish pink & 199 & 163 & 185 \\
78 & Dark yellowish brown & 80 & 52 & 26 & 169 & Strong greenish blue & 19 & 133 & 175 & 254 & Vivid purplish red & 221 & 35 & 136 \\
79 & Light grayish yellowish brown & 180 & 155 & 141 & 171 & Very light greenish blue & 148 & 214 & 239 & 255 & Strong purplish red & 184 & 55 & 115 \\
80 & Grayish yellowish brown & 126 & 105 & 93 & 172 & Light greenish blue & 101 & 168 & 195 & 256 & Deep purplish red & 136 & 16 & 85 \\
81 & Dark grayish yellowish brown & 77 & 61 & 51 & 173 & Moderate greenish blue & 42 & 118 & 145 & 257 & Very deep purplish red & 84 & 6 & 60 \\
82 & Vivid yellow & 241 & 191 & 21 & 174 & Dark greenish blue & 19 & 74 & 96 & 258 & Moderate purplish red & 171 & 75 & 116 \\
83 & Brilliant yellow & 247 & 206 & 80 & 175 & Very dark greenish blue & 11 & 44 & 59 & 259 & Dark purplish red & 110 & 41 & 76 \\
84 & Strong yellow & 217 & 174 & 47 & 176 & Vivid blue & 27 & 92 & 215 & 260 & Very dark purplish red & 67 & 20 & 50 \\
85 & Deep yellow & 184 & 143 & 22 & 177 & Brilliant blue & 65 & 157 & 237 & 261 & Light grayish purplish red & 178 & 135 & 155 \\
86 & Light yellow & 244 & 210 & 132 & 178 & Strong blue & 39 & 108 & 189 & 262 & Grayish purplish red & 148 & 92 & 115 \\
87 & Moderate yellow & 210 & 175 & 99 &  &  &  &  &  & 263 & White & 231 & 225 & 233 \\
88 & Dark yellow & 176 & 143 & 66 &  &  &  &  &  & 264 & Light gray & 189 & 183 & 191 \\
89 & Pale yellow & 239 & 215 & 178 &  &  &  &  &  & 265 & Medium gray & 138 & 132 & 137 \\
 &  &  &  &  &  &  &  &  &  & 266 & Dark gray & 88 & 84 & 88 \\
 &  &  &  &  &  &  &  &  &  & 267 & Black & 43 & 41 & 43 \\
\hline
\end{tabular}
}
\caption{List of ISCC NBS Level 3 Colors used in the GenColorBench evaluation}
\label{tab:l3_colors}
\end{table*}
\begin{table*}[!t]
\centering
\resizebox{\textwidth}{!}{
\begin{tabular}{|c|l|c|c|c|c||c|l|c|c|c|c||c|l|c|c|c|c|}
\hline
\textbf{ID} & \textbf{Color Name} & \textbf\{hex\} & \textbf{R} & \textbf{G} & \textbf{B} & \textbf{ID} & \textbf{Color Name} & \textbf\{hex\} & \textbf{R} & \textbf{G} & \textbf{B} & \textbf{ID} & \textbf{Color Name} & \textbf\{hex\} & \textbf{R} & \textbf{G} & \textbf{B} \\
\hline
1  & AliceBlue            & \#F0F8FF & 240 & 248 & 255 & 50 & LightBlue            & \#ADD8E6 & 173 & 216 & 230 & 99  & PowderBlue           & \#B0E0E6 & 176 & 224 & 230 \\
2  & AntiqueWhite         & \#FAEBD7 & 250 & 235 & 215 & 51 & LightCoral           & \#F08080 & 240 & 128 & 128 & 100 & Purple               & \#800080 & 128 & 0   & 128 \\
3  & Aqua                 & \#00FFFF & 0   & 255 & 255 & 52 & LightCyan            & \#E0FFFF & 224 & 255 & 255 & 101 & RebeccaPurple        & \#663399 & 102 & 51  & 153 \\
4  & Aquamarine           & \#7FFFD4 & 127 & 255 & 212 & 53 & LightGoldenRodYellow & \#FAFAD2 & 250 & 250 & 210 & 102 & Red                  & \#FF0000 & 255 & 0   & 0   \\
5  & Azure                & \#F0FFFF & 240 & 255 & 255 & 54 & LightGray            & \#D3D3D3 & 211 & 211 & 211 & 103 & RosyBrown            & \#BC8F8F & 188 & 143 & 143 \\
6  & Beige                & \#F5F5DC & 245 & 245 & 220 & 55 & LightGrey            & \#D3D3D3 & 211 & 211 & 211 & 104 & RoyalBlue            & \#4169E1 & 65  & 105 & 225 \\
7  & Bisque               & \#FFE4C4 & 255 & 228 & 196 & 56 & LightGreen           & \#90EE90 & 144 & 238 & 144 & 105 & SaddleBrown          & \#8B4513 & 139 & 69  & 19  \\
8  & Black                & \#000000 & 0   & 0   & 0   & 57 & LightPink            & \#FFB6C1 & 255 & 182 & 193 & 106 & Salmon               & \#FA8072 & 250 & 128 & 114 \\
9  & BlanchedAlmond       & \#FFEBCD & 255 & 235 & 205 & 58 & LightSalmon          & \#FFA07A & 255 & 160 & 122 & 107 & SandyBrown           & \#F4A460 & 244 & 164 & 96  \\
10 & Blue                 & \#0000FF & 0   & 0   & 255 & 59 & LightSeaGreen        & \#20B2AA & 32  & 178 & 170 & 108 & SeaGreen             & \#2E8B57 & 46  & 139 & 87  \\
11 & BlueViolet           & \#8A2BE2 & 138 & 43  & 226 & 60 & LightSkyBlue         & \#87CEFA & 135 & 206 & 250 & 109 & SeaShell             & \#FFF5EE & 255 & 245 & 238 \\
12 & Brown                & \#A52A2A & 165 & 42  & 42  & 61 & LightSlateGray       & \#778899 & 119 & 136 & 153 & 110 & Sienna               & \#A0522D & 160 & 82  & 45  \\
13 & BurlyWood            & \#DEB887 & 222 & 184 & 135 & 62 & LightSlateGrey       & \#778899 & 119 & 136 & 153 & 111 & Silver               & \#C0C0C0 & 192 & 192 & 192 \\
14 & CadetBlue            & \#5F9EA0 & 95  & 158 & 160 & 63 & LightSteelBlue       & \#B0C4DE & 176 & 196 & 222 & 112 & SkyBlue              & \#87CEEB & 135 & 206 & 235 \\
15 & Chartreuse           & \#7FFF00 & 127 & 255 & 0   & 64 & LightYellow          & \#FFFFE0 & 255 & 255 & 224 & 113 & SlateBlue            & \#6A5ACD & 106 & 90  & 205 \\
16 & Chocolate            & \#D2691E & 210 & 105 & 30  & 65 & Lime                 & \#00FF00 & 0   & 255 & 0   & 114 & SlateGray            & \#708090 & 112 & 128 & 144 \\
17 & Coral                & \#FF7F50 & 255 & 127 & 80  & 66 & LimeGreen            & \#32CD32 & 50  & 205 & 50  & 115 & SlateGrey            & \#708090 & 112 & 128 & 144 \\
18 & CornflowerBlue       & \#6495ED & 100 & 149 & 237 & 67 & Linen                & \#FAF0E6 & 250 & 240 & 230 & 116 & Snow                 & \#FFFAFA & 255 & 250 & 250 \\
19 & Cornsilk             & \#FFF8DC & 255 & 248 & 220 & 68 & Magenta              & \#FF00FF & 255 & 0   & 255 & 117 & SpringGreen          & \#00FF7F & 0   & 255 & 127 \\
20 & Crimson              & \#DC143C & 220 & 20  & 60  & 69 & Maroon               & \#800000 & 128 & 0   & 0   & 118 & SteelBlue            & \#4682B4 & 70  & 130 & 180 \\
21 & Cyan                 & \#00FFFF & 0   & 255 & 255 & 70 & MediumAquaMarine     & \#66CDAA & 102 & 205 & 170 & 119 & Tan                  & \#D2B48C & 210 & 180 & 140 \\
22 & DarkBlue             & \#00008B & 0   & 0   & 139 & 71 & MediumBlue           & \#0000CD & 0   & 0   & 205 & 120 & Teal                 & \#008080 & 0   & 128 & 128 \\
23 & DarkCyan             & \#008B8B & 0   & 139 & 139 & 72 & MediumOrchid         & \#BA55D3 & 186 & 85  & 211 & 121 & Thistle              & \#D8BFD8 & 216 & 191 & 216 \\
24 & DarkGoldenRod        & \#B8860B & 184 & 134 & 11  & 73 & MediumPurple         & \#9370DB & 147 & 112 & 219 & 122 & Tomato               & \#FF6347 & 255 & 99  & 71  \\
25 & DarkGray             & \#A9A9A9 & 169 & 169 & 169 & 74 & MediumSeaGreen       & \#3CB371 & 60  & 179 & 113 & 123 & Turquoise            & \#40E0D0 & 64  & 224 & 208 \\
26 & DarkGrey             & \#A9A9A9 & 169 & 169 & 169 & 75 & MediumSlateBlue      & \#7B68EE & 123 & 104 & 238 & 124 & Violet               & \#EE82EE & 238 & 130 & 238 \\
27 & DarkGreen            & \#006400 & 0   & 100 & 0   & 76 & MediumSpringGreen    & \#00FA9A & 0   & 250 & 154 & 125 & Wheat                & \#F5DEB3 & 245 & 222 & 179 \\
28 & DarkKhaki            & \#BDB76B & 189 & 183 & 107 & 77 & MediumTurquoise      & \#48D1CC & 72  & 209 & 204 & 126 & White                & \#FFFFFF & 255 & 255 & 255 \\
29 & DarkMagenta          & \#8B008B & 139 & 0   & 139 & 78 & MediumVioletRed      & \#C71585 & 199 & 21  & 133 & 127 & WhiteSmoke           & \#F5F5F5 & 245 & 245 & 245 \\
30 & DarkOliveGreen       & \#556B2F & 85  & 107 & 47  & 79 & MidnightBlue         & \#191970 & 25  & 25  & 112 & 128 & Yellow               & \#FFFF00 & 255 & 255 & 0   \\
31 & DarkOrange           & \#FF8C00 & 255 & 140 & 0   & 80 & MintCream            & \#F5FFFA & 245 & 255 & 250 & 129 & YellowGreen          & \#9ACD32 & 154 & 205 & 50  \\
32 & DarkOrchid           & \#9932CC & 153 & 50  & 204 & 81 & MistyRose            & \#FFE4E1 & 255 & 228 & 225 &     &                      &          &     &     &     \\
33 & DarkRed              & \#8B0000 & 139 & 0   & 0   & 82 & Moccasin             & \#FFE4B5 & 255 & 228 & 181 &     &                      &          &     &     &     \\
34 & DarkSalmon           & \#E9967A & 233 & 150 & 122 & 83 & NavajoWhite          & \#FFDEAD & 255 & 222 & 173 &     &                      &          &     &     &     \\
35 & DarkSeaGreen         & \#8FBC8F & 143 & 188 & 143 & 84 & Navy                 & \#000080 & 0   & 0   & 128 &     &                      &          &     &     &     \\
36 & DarkSlateBlue        & \#483D8B & 72  & 61  & 139 & 85 & OldLace              & \#FDF5E6 & 253 & 245 & 230 &     &                      &          &     &     &     \\
37 & DarkSlateGray        & \#2F4F4F & 47  & 79  & 79  & 86 & Olive                & \#808000 & 128 & 128 & 0   &     &                      &          &     &     &     \\
38 & DarkSlateGrey        & \#2F4F4F & 47  & 79  & 79  & 87 & OliveDrab            & \#6B8E23 & 107 & 142 & 35  &     &                      &          &     &     &     \\
39 & DarkTurquoise        & \#00CED1 & 0   & 206 & 209 & 88 & Orange               & \#FFA500 & 255 & 165 & 0   &     &                      &          &     &     &     \\
40 & DarkViolet           & \#9400D3 & 148 & 0   & 211 & 89 & OrangeRed            & \#FF4500 & 255 & 69  & 0   &     &                      &          &     &     &     \\
41 & DeepPink             & \#FF1493 & 255 & 20  & 147 & 90 & Orchid               & \#DA70D6 & 218 & 112 & 214 &     &                      &          &     &     &     \\
42 & DeepSkyBlue          & \#00BFFF & 0   & 191 & 255 & 91 & PaleGoldenRod        & \#EEE8AA & 238 & 232 & 170 &     &                      &          &     &     &     \\
43 & DimGray              & \#696969 & 105 & 105 & 105 & 92 & PaleGreen            & \#98FB98 & 152 & 251 & 152 &     &                      &          &     &     &     \\
44 & DimGrey              & \#696969 & 105 & 105 & 105 & 93 & PaleTurquoise        & \#AFEEEE & 175 & 238 & 238 &     &                      &          &     &     &     \\
45 & DodgerBlue           & \#1E90FF & 30  & 144 & 255 & 94 & PaleVioletRed        & \#DB7093 & 219 & 112 & 147 &     &                      &          &     &     &     \\
46 & FireBrick            & \#B22222 & 178 & 34  & 34  & 95 & PapayaWhip           & \#FFEFD5 & 255 & 239 & 213 &     &                      &          &     &     &     \\
47 & FloralWhite          & \#FFFAF0 & 255 & 250 & 240 & 96 & PeachPuff            & \#FFDAB9 & 255 & 218 & 185 &     &                      &          &     &     &     \\
48 & ForestGreen          & \#228B22 & 34  & 139 & 34  & 97 & Peru                 & \#CD853F & 205 & 133 & 63  &     &                      &          &     &     &     \\
49 & Fuchsia              & \#FF00FF & 255 & 0   & 255 & 98 & Pink                 & \#FFC0CB & 255 & 192 & 203 &     &                      &          &     &     &     \\
\hline
\end{tabular}
}
\caption{List of CSS3/X11 colors used in GenColorBench evaluation.}
\label{tab:css_colors}
\end{table*}

\section{Object Categorization}

To ensure a thorough evaluation of color generation in text-to-image (T2I) generation, we curate a set of 108 diverse objects that span a wide range of visual and semantic categories. This object set allows us to test how well generative models handle color prompts across different shapes, materials, and contexts. We draw these objects from two well-established datasets: COCO\citep{lin2014microsoft} and ImageNet\citep{deng2009imagenet}, both of which offer a large pool of visually distinct and commonly recognized objects. Each object is selected with careful consideration of three criteria: (1) recognizability in T2I generation, ensuring that current models can reliably render the object given a prompt (e.g., "a blue chair"), (2) plausible color variability, allowing the object to appear in a wide range of colors (e.g., a shirt or a car), and (3) segmentation suitability, so the object can be cleanly separated from its background using segmentation models.

While conducting initial experiments, we observed that many generated images include additional visual components associated with the main object, but which are not relevant for color evaluation. For example, when prompting for a "red car," the generated image may include elements such as tires, headlights, or windows—components that differ in material and expected color from the car’s painted body. Including these in the mask during evaluation would introduce noise and bias in the color measurements. To address this issue, we introduce the concept of negative labels---subcomponents or associated elements of an object that should be excluded from the evaluation mask. For each object class, we generate a list of such negative labels using GPT-4o, leveraging its broad world knowledge and language understanding to identify parts that are typically not relevant for color fidelity. These negative labels help refine the segmentation masks, allowing us to more precisely isolate the region of interest (e.g., the body of a car), and ensure a fair and focused color assessment. All the objects along with their negative labels are listed in Table~\ref{tab:dataset_classification_coco}, and Table~\ref{tab:dataset_classification_imagenet}.

\begin{table*}[!ht]
\centering
\resizebox{\textwidth}{!}{
\begin{tabular}{|l|l|l|p{15cm}|}
\hline
\textbf{Dataset} & \textbf{Category} & \textbf{Class Name} & \textbf{Negative Labels} \\
\hline
COCO & Vehicle & vehicle & glass windows. front glass windows. back glass windows. windshield. black wheel. tire. metal rim. headlight. taillight. mirror. metal bumper. license plate. roof rack. antenna. grille. door handle. door frame. side mirror glass. hubcap. mudguard. windshield wiper. \\
\hline
COCO & Vehicle & bicycle & wheel. black tire. rim. seat. basket. pedal. steel handle. metal handle. metal nut. chain. brake disc. reflector. gear shift. handlebar grip. spoke. \\
\hline
COCO & Vehicle & car & window. windshield. wheel. tire. rim. headlight. taillight. mirror. bumper. license plate. roof rack. antenna. grille. door handle. door frame. side mirror glass. hubcap. mudguard. windshield wiper. exhaust pipe. \\
\hline
COCO & Vehicle & motorcycle & black wheel. tire. engine. rim. headlight. taillight. mirror. exhaust. leather seat. handlebar. chain. foot peg. black handle. metal stand. brake lever. turn signal. \\
\hline
COCO & Vehicle & airplane & window. glass windows. front glass window. back glass window. plane engine. black tire. antenna. tail fin. wing flap. landing gear. cockpit window. propeller. \\
\hline
COCO & Vehicle & bus & window. front glass windows. back glass window. windshield. black wheel. tire. rim. headlight. taillight. mirror. bumper. license plate. door handle. roof rack. antenna. side mirror glass. exhaust pipe. stop sign. \\
\hline
COCO & Vehicle & train & window. door. wheel. bogie. pantograph. headlight. buffer. coupling. antenna. rail. overhead line. windshield. wiper. light. connector. ladder. handle. logo. text. number. symbol. pipe. cable. horn. vent. exhaust. panel. joint. frame. suspension. track. gravel. pole. wire. signboard. indicator. sticker. emblem. \\
\hline
COCO & Vehicle & truck & window. windshield. wheel. tire. rim. headlight. taillight. mirror. bumper. license plate. roof rack. antenna. grille. door handle. exhaust pipe. side mirror glass. mudflap. \\
\hline
COCO & Vehicle & boat & window. hull detail. motor. propeller. antenna. railing. rope. flag. deck equipment. lifebuoy. mast. anchor. \\
\hline
COCO & Fruit and Veg & banana & sticker. wrap. stem. leaf. spot. bruise. brown spot. peel. string. \\
\hline
COCO & Fruit and Veg & apple & sticker. wrap. stem. leaf. spot. bruise. stem scar. wax coating. calyx. \\
\hline
COCO & Fruit and Veg & orange & sticker. wrap. stem. leaf. spot. bruise. peel. segment. pith. \\
\hline
COCO & Fruit and Veg & broccoli & stem. leaf. spot. bruise. floret. stalk. \\
\hline
COCO & Fruit and Veg & carrot & stem. leaf. spot. dirt. root tip. soil. \\
\hline
COCO & furniture and Household & chair & cushion. leg. screw. joint. tag. upholstery. armrest. backrest. \\
\hline
COCO & furniture and Household & couch & cushion. leg. seam. button. pillow. tag. upholstery. armrest. backrest. \\
\hline
COCO & furniture and Household & potted plant & pot. soil. label. stick. tag. drainage hole. saucer. \\
\hline
COCO & furniture and Household & sink & faucet only. drain. soap dispenser. knob. handle above sink. countertop. \\
\hline
COCO & furniture and Household & book & cover. spine. bookmark. sticker. dust jacket. pages. \\
\hline
COCO & furniture and Household & clock & glass. hands. dial. numbers. knob. frame. pendulum. \\
\hline
COCO & furniture and Household & vase & rim. neck. base. chip. pattern. glaze. \\
\hline
COCO & animals & cat & eyes. teeth. nose. mouth. tongue. paw. claw. ear. whiskers. tail. fur. collar. leash. tag. accessory. fur pattern. fur texture. \\
\hline
COCO & animals & dog & eyes. teeth. nose. mouth. tongue. paw. claw. ear. whiskers. tail. fur. collar. leash. tag. accessory. fur pattern. fur texture. \\
\hline
COCO & animals & horse & eyes. teeth. nose. mouth. tongue. hoof. claw. ear. whiskers. tail. mane. fur. saddle. reins. harness. bridle. coat pattern. \\
\hline
COCO & animals & sheep & eyes. teeth. nose. mouth. tongue. hoof. claw. ear. whiskers. tail. wool. tag. collar. fleece texture. \\
\hline
COCO & animals & cow & eyes. teeth. nose. mouth. tongue. hoof. claw. ear. whiskers. tail. fur. tag. bell. collar. coat pattern. \\
\hline
COCO & animals & elephant & eyes. teeth. nose. mouth. tongue. paw. claw. ear. whiskers. tail. tusk. mane. fur. chain. tusk cover. cloth. skin folds. \\
\hline
COCO & animals & bear & eyes. teeth. nose. mouth. tongue. paw. claw. ear. whiskers. tail. fur. collar. accessory. fur pattern. \\
\hline
COCO & animals & zebra & eyes. teeth. nose. mouth. tongue. paw. claw. ear. whiskers. tail. stripes. fur. harness. accessory. \\
\hline
COCO & animals & giraffe & eyes. teeth. nose. mouth. tongue. paw. claw. ear. whiskers. tail. fur. tag. collar. spots. \\
\hline
COCO & cloths and Accessories & tie & text. logo. monogram. tag. knot. \\
\hline
COCO & cloths and Accessories & handbag & text. logo. monogram. inner layer. tag. seam. button. zipper. strap. \\
\hline
COCO & cloths and Accessories & backpack & text. logo. monogram. inner layer. tag. seam. button. zipper. strap. buckle. pocket. \\
\hline
COCO & cloths and Accessories & suitcase & text. logo. monogram. inner layer. tag. seam. button. zipper. handle. wheel. lock. zipper pull. \\
\hline
COCO & cloths and Accessories & umbrella & text. logo. monogram. tag. seam. pipe. handle. fabric folds. \\
\hline
COCO & sports and toys & sports ball & logo. text. stitching. brand name. grip tape. valve. panel seams. \\
\hline
COCO & sports and toys & baseball bat & logo. text. grip tape. brand name. handle. barrel. \\
\hline
COCO & sports and toys & kite & string. tail. frame. handle. spars. \\
\hline
COCO & sports and toys & frisbee & logo. text. brand name. rim. \\
\hline
COCO & sports and toys & surfboard & logo. text. brand name. leash. fin. deck pad. \\
\hline
COCO & sports and toys & skis & logo. text. brand name. binding. tip. tail. \\
\hline
COCO & sports and toys & baseball glove & logo. text. stitching. brand name. webbing. \\
\hline
COCO & sports and toys & skateboard & wheel. logo. text. grip tape. brand name. trucks. deck. \\
\hline
COCO & Tools and Misc & hair drier & button. switch. cord. label. nozzle. \\
\hline
COCO & Tools and Misc & remote & button. label. logo. screen. \\
\hline
COCO & Tools and Misc & microwave & button. label. logo. handle. display panel. \\
\hline
COCO & Tools and Misc & sink & faucet. drain. knob. handle. basin. \\
\hline
COCO & Tools and Misc & toaster & button. label. logo. slot. \\
\hline
COCO & Tools and Misc & refrigerator & handle. logo. label. button. door seal. \\
\hline
COCO & Tools and Misc & oven & handle. button. label. logo. control panel. \\
\hline
COCO & Tools and Misc & knife & handle. logo. brand name. blade edge. \\
\hline
\end{tabular}
}
\caption{List of the objects selected from the COCO dataset.}
\label{tab:dataset_classification_coco}
\end{table*}
\begin{table*}[!ht]
\centering
\resizebox{\textwidth}{!}{
\begin{tabular}{|l|l|l|p{15cm}|}
\hline
\textbf{Dataset} & \textbf{Category} & \textbf{Class Name} & \textbf{Negative Labels} \\
\hline
ImageNet & Vehicle & ambulance & window. windshield. wheel. tire. rim. headlight. taillight. mirror. bumper. license plate. roof rack. antenna. grille. door handle. siren. \\
\hline
ImageNet & Vehicle & beach wagon & wheel. tire. rim. seat. handlebar. pedal. chain. brake. reflector. bell. gear. \\
\hline
ImageNet & Vehicle & jeep & window. windshield. wheel. tire. rim. headlight. taillight. mirror. bumper. license plate. roof rack. antenna. grille. door handle. \\
\hline
ImageNet & Vehicle & minivan & window. windshield. wheel. tire. rim. headlight. taillight. mirror. bumper. license plate. roof rack. antenna. grille. door handle. \\
\hline
ImageNet & Vehicle & sports car & window. windshield. wheel. tire. rim. headlight. taillight. mirror. bumper. license plate. roof rack. antenna. grille. door handle. \\
\hline
ImageNet & Vehicle & tow truck & window. windshield. wheel. tire. rim. headlight. taillight. mirror. bumper. license plate. roof rack. antenna. grille. door handle. tow hook. \\
\hline
ImageNet & Vehicle & ferry & window. hull detail. motor. propeller. antenna. railing. rope. flag. deck equipment. lifeboat. \\
\hline
ImageNet & Vehicle & taxi & window. windshield. wheel. tire. rim. headlight. taillight. mirror. bumper. license plate. roof rack. antenna. grille. door handle. taxi sign. \\
\hline
ImageNet & Fruit and Veg & lemon & sticker. wrap. stem. leaf. spot. bruise. peel texture. \\
\hline
ImageNet & Fruit and Veg & mango & sticker. wrap. stem. leaf. spot. bruise. peel texture. \\
\hline
ImageNet & Fruit and Veg & papaya & sticker. wrap. stem. leaf. spot. bruise. seeds. peel texture. \\
\hline
ImageNet & Fruit and Veg & guava & sticker. wrap. stem. leaf. spot. bruise. seeds. peel texture. \\
\hline
ImageNet & Fruit and Veg & strawberry & sticker. wrap. stem. leaf. spot. bruise. seeds. calyx. \\
\hline
ImageNet & furniture and Household & teapot & lid. handle. spout. base. knob. \\
\hline
ImageNet & furniture and Household & table & leg. joint. screw. tabletop. \\
\hline
ImageNet & furniture and Household & desk & leg. joint. screw. drawer. \\
\hline
ImageNet & furniture and Household & bookcase & shelf. joint. screw. back panel. \\
\hline
ImageNet & furniture and Household & wardrobe & handle. knob. hinge. door. \\
\hline
ImageNet & furniture and Household & mug & handle. rim. base. \\
\hline
ImageNet & furniture and Household & candle & wick. flame. holder. wax. \\
\hline
ImageNet & animals & tiger & eyes. teeth. nose. mouth. tongue. paw. claw. ear. whiskers. tail. stripes. fur. mane. \\
\hline
ImageNet & animals & parrot & beak. claw. wingtip. feather. cage. perch. tag. \\
\hline
ImageNet & animals & duck & beak. claw. wingtip. feather. tag. accessory. \\
\hline
ImageNet & animals & crocodile & eyes. teeth. scales. tail. claw. \\
\hline
ImageNet & animals & shark & eyes. teeth. fin. tail. gills. \\
\hline
ImageNet & animals & lobster & claw band. eyes. antenna. legs. shell. \\
\hline
ImageNet & animals & goldfish & eyes. fins. tail. bowl. accessory. \\
\hline
ImageNet & animals & turtle & eyes. shell. legs. tail. scales. \\
\hline
ImageNet & animals & owl & eyes. beak. wingtip. feather. talons. \\
\hline
ImageNet & cloths and Accessories & T-shirt & text. logo. monogram. inner layer. tag. seam. button. collar. cuff. pocket. \\
\hline
ImageNet & cloths and Accessories & sweatshirt & text. logo. monogram. inner layer. tag. seam. button. collar. cuff. pocket. \\
\hline
ImageNet & cloths and Accessories & suit & text. logo. monogram. inner layer. tag. seam. button. collar. cuff. pocket. \\
\hline
ImageNet & cloths and Accessories & jacket & text. logo. monogram. inner layer. tag. seam. button. collar. cuff. pocket. \\
\hline
ImageNet & cloths and Accessories & coat & text. logo. monogram. inner layer. tag. seam. button. collar. cuff. pocket. \\
\hline
ImageNet & cloths and Accessories & jeans & text. logo. monogram. inner layer. tag. seam. button. pocket. \\
\hline
ImageNet & cloths and Accessories & pants & text. logo. monogram. inner layer. tag. seam. button. pocket. \\
\hline
ImageNet & cloths and Accessories & shorts & text. logo. monogram. inner layer. tag. seam. button. pocket. \\
\hline
ImageNet & cloths and Accessories & hat & text. logo. monogram. inner layer. tag. seam. button. \\
\hline
ImageNet & sports and toys & football helmet & logo. text. stitching. brand name. padding. strap. face guard. \\
\hline
ImageNet & sports and toys & golf ball & logo. text. stitching. brand name. dimples. \\
\hline
ImageNet & sports and toys & boxing glove & logo. text. stitching. brand name. strap. \\
\hline
ImageNet & sports and toys & teddy bear & eyes. nose. mouth. paw. claw. ear. tongue. collar. tag. accessory. fur. stitching. \\
\hline
ImageNet & sports and toys & snowboard & logo. text. brand name. leash. binding. fin. \\
\hline
ImageNet & sports and toys & balloon & string. knot. valve. \\
\hline
ImageNet & sports and toys & doll & string. hair. eyes. mouth. nose. limbs. dress. accessory. tag. \\
\hline
ImageNet & sports and toys & toy poodle & eyes. nose. mouth. paw. claw. ear. tongue. collar. tag. accessory. fur. stitching. \\
\hline
ImageNet & sports and toys & toy terrier & eyes. nose. mouth. paw. claw. ear. tongue. collar. tag. accessory. fur. stitching. \\
\hline
ImageNet & Tools and Misc & sponge & label. tag. pores. \\
\hline
ImageNet & Tools and Misc & cutting board & knife marks. logo. text. \\
\hline
ImageNet & Tools and Misc & computer mouse & button. logo. cable. scroll wheel. \\
\hline
ImageNet & Tools and Misc & hair dryer & button. switch. cord. label. nozzle. \\
\hline
ImageNet & Tools and Misc & iron & button. switch. cord. label. soleplate. \\
\hline
ImageNet & Tools and Misc & fan & button. switch. cord. label. blades. \\
\hline
ImageNet & Tools and Misc & hammer & handle. brand name. logo. claw. \\
\hline
ImageNet & Tools and Misc & wrench & handle. brand name. logo. jaw. \\
\hline
ImageNet & Tools and Misc & saw & handle. brand name. logo. blade. \\
\hline
ImageNet & Tools and Misc & ruler & markings. text. logo. edges. \\
\hline
\end{tabular}
}
\caption{List of the objects selected from the ImageNet dataset.}
\label{tab:dataset_classification_imagenet}
\end{table*}

\section{Prompt Templates}
We organize prompts into four levels of difficulty, based on their semantic and compositional complexity: (1) Object-Focused Prompts---These are straightforward, object-centric prompts that mention only a single object and its target color (e.g., “a red apple” or “a green chair”). The goal here is to assess basic color understanding, including both color name fidelity and numerical color consistency with the expected RGB/hex value. These prompts are the most direct and least ambiguous prompts.
(2) Contextual Prompts---the object is described within a broader scene or setting, but only a single colored object is mentioned (e.g., “a blue vase on a wooden table near a window”). These prompts evaluate the model’s ability to preserve the intended color within contextual descriptions, measuring both color name accuracy and the association of the correct color to the correct object in a scene. (3) Scene Descriptive Prompts---These prompts describe scenes containing multiple objects, each associated with its own color (e.g., “a red apple next to a green pear and a yellow banana”). This level tests the model’s ability to distinguish and correctly apply multiple colors to multiple objects, and is useful for evaluating multi-object compositionality and the avoidance of color-object entanglement. (4) Implicit Color Association---These are the most semantically complex prompts, involving color references between objects (e.g., “a cup that is the same color as the nearby blue notebook”, or “a cat lying on a rug that shares its pink color”). Here, only one object is explicitly assigned a color, and the second object's color is described relationally. These prompts assess whether models can understand and generate color consistency through indirect, reference-based language. We list all the prompt templates below.

\subsection{List of Object Focused Prompt Templates.}

\begin{enumerate}
    \item A \{color\} \{object\}
    \item The \{object\} is \{color\}
    \item A photo of a \{color\} \{object\}
    \item A \{object\} that is entirely \{color\}
    \item An image of a \{color\} \{object\}
    \item A \{color\} colored \{object\}
    \item A single \{color\} \{object\}
    \item A \{object\}, and it’s \{color\}
    \item A \{object\} in a \{color\} color
    \item A \{object\} rendered in \{color\} color
    \item A \{object\} with a \{color\} color
    \item A realistic \{object\} in \{color\}
    \item An image of a \{object\} in hex color \{hex\}
    \item A \{object\} in color \{hex\}
    \item A \{object\} with hex color \{hex\}
    \item A close-up of a \{object\} in the color \{hex\}
    \item A \{object\} rendered in \{hex\} color
    \item A photo of a \{object\} in the color \{hex\}
    \item A \{object\} rendered entirely in \{hex\}
    \item A \{object\} designed in \{hex\} color
    \item A realistic \{hex\}-colored \{object\}
    \item A highly detailed \{object\} in hex \{hex\}
    \item A \{object\} in rgb(\{r\}, \{g\}, \{b\})
    \item A \{object\} with the color rgb(\{r\}, \{g\}, \{b\})
    \item A \{object\} rendered in RGB color rgb(\{r\}, \{g\}, \{b\})
    \item A photo of a \{object\} in color rgb(\{r\}, \{g\}, \{b\})
    \item A \{object\} with color rgb(\{r\}, \{g\}, \{b\})
\end{enumerate}

\subsection{List of Contextual Prompt Templates.}
\begin{enumerate}
    \item A \{color\} apple on a white plate
    \item A \{color\} banana next to a sliced orange
    \item A \{color\} carrot placed on a kitchen counter
    \item A \{color\} mango in a fruit bowl with a lemon
    \item A \{color\} strawberry on top of a dessert plate
    \item A \{color\} broccoli beside a cutting board
    \item A \{color\} guava resting in a wire fruit basket
    \item A \{color\} papaya cut in half on a wooden table
    \item A \{color\} lemon on a breakfast tray
    \item A \{color\} car parked near a sidewalk
    \item A \{color\} truck beside a loading dock
    \item A \{color\} bus at a bus stop
    \item A \{color\} motorcycle on a street corner
    \item A \{color\} taxi in front of a building
    \item A \{color\} jeep driving along a dirt road
    \item A \{color\} sports car on a highway
    \item A \{color\} train at a rural station
    \item A \{color\} ferry approaching the dock
    \item A \{color\} airplane at the runway gate
    \item A \{color\} chair next to a wooden table
    \item A \{color\} couch in front of a window
    \item A \{color\} potted plant on a bookshelf
    \item A \{color\} teapot on a breakfast tray
    \item A \{color\} clock on a white wall
    \item A \{color\} vase placed on a dining table
    \item A \{color\} mug on a desk with books
    \item A \{color\} candle beside a mirror
    \item A \{color\} wardrobe beside a small chair
    \item A \{color\} sink installed in a marble countertop
    \item A \{color\} cat sleeping on a couch
    \item A \{color\} dog playing with a ball
    \item A \{color\} horse standing in a stable
    \item A \{color\} sheep grazing on a green field
    \item A \{color\} cow near a wooden fence
    \item A \{color\} tiger behind a jungle bush
    \item A \{color\} parrot on a tree branch
    \item A \{color\} duck floating on a pond
    \item A \{color\} owl perched on a wooden stump
    \item A \{color\} goldfish swimming in a small tank
    \item A \{color\} T-shirt folded on a table
    \item A \{color\} jacket hanging on a coat rack
    \item A \{color\} pair of jeans on a bed
    \item A \{color\} hat resting on a chair
    \item A \{color\} tie draped over a hanger
    \item A \{color\} coat hanging near the door
    \item A \{color\} backpack leaning against the wall
    \item A \{color\} handbag on a desk
    \item A \{color\} sports ball on a gym floor
    \item A \{color\} kite flying in a clear sky
    \item A \{color\} baseball glove on a bench
    \item A \{color\} frisbee lying on the grass
    \item A \{color\} snowboard resting against a wall
    \item A \{color\} teddy bear on a child's bed
    \item A \{color\} boxing glove placed on a shelf
    \item A \{color\} doll sitting in a toy stroller
    \item A \{color\} microwave on a kitchen shelf
    \item A \{color\} hair dryer on a bathroom counter
    \item A \{color\} toaster beside a coffee machine
    \item A \{color\} refrigerator in the corner of the kitchen
    \item A \{color\} cutting board on a kitchen island
    \item A \{color\} sponge near a faucet
    \item A \{color\} ruler beside an open notebook
    \item A \{color\} fan placed near a window
\end{enumerate}

\subsection{List of Scene Descriptive Prompt Templates.}

\begin{enumerate}
\item A \{color1\} banana and a \{color2\} apple on a wooden table
\item A \{color1\} dog next to a \{color2\} cat and a \{color3\} couch in a living room
\item A \{color1\} skateboard beside a \{color2\} sports ball and a \{color3\} baseball bat
\item A \{color1\} jeep parked near a \{color2\} ambulance on a rainy street
\item A \{color1\} T-shirt and a \{color2\} pair of jeans folded on a bed
\item A \{color1\} microwave next to a \{color2\} refrigerator and a \{color3\} toaster
\item A \{color1\} tie hanging beside a \{color2\} hat and a \{color3\} jacket
\item A \{color1\} zebra standing with a \{color2\} giraffe in the savannah
\item A \{color1\} teddy bear and a \{color2\} doll placed on a shelf
\item A \{color1\} papaya, a \{color2\} guava, and a \{color3\} lemon in a fruit basket
\item A \{color1\} goldfish swimming with a \{color2\} turtle and a \{color3\} shark
\item A \{color1\} chair and a \{color2\} desk in a sunlit room
\item A \{color1\} surfboard, a \{color2\} kite, and a \{color3\} frisbee on the beach
\item A \{color1\} oven and a \{color2\} sink in a small kitchen
\item A \{color1\} cow grazing with a \{color2\} horse in a green field
\item A \{color1\} boat and a \{color2\} ferry docked at the harbor
\item A \{color1\} wardrobe and a \{color2\} bookcase against a blue wall
\item A \{color1\} duck floating near a \{color2\} parrot perched on a tree
\item A \{color1\} fan, a \{color2\} computer mouse, and a \{color3\} cutting board on the table
\item A \{color1\} umbrella leaning against a \{color2\} suitcase
\item A \{color1\} couch with a \{color2\} potted plant beside it
\item A \{color1\} car parked near a \{color2\} bus and a \{color3\} truck
\item A \{color1\} bear standing near a \{color2\} elephant in the wild
\item A \{color1\} book and a \{color2\} clock on a wooden shelf
\item A \{color1\} balloon tied to a \{color2\} snowboard and a \{color3\} boxing glove
\item A \{color1\} sink and a \{color2\} hair dryer on a bathroom counter
\item A \{color1\} strawberry and a \{color2\} mango next to a \{color3\} orange
\item A \{color1\} tie and a \{color2\} backpack on a desk
\item A \{color1\} shark chasing a \{color2\} lobster underwater
\item A \{color1\} remote beside a \{color2\} mug and a \{color3\} candle
\end{enumerate}

\subsection{Implicit Color Association Prompt Templates}
\begin{enumerate}
\item A \{color\} backpack is placed next to a suitcase that has the same color as the backpack, making their colors clearly match
\item A bicycle painted in \{color\} is parked beside a car that shares this exact color, making their similarity obvious
A \{color\} dog is sitting near a cat whose fur matches the dog’s color perfectly.
\item A \{color\} chair is positioned close to a couch that is painted in the same color, showing clear color similarity
\item An airplane painted \{color\} flies above a bus that is painted the same color, making the matching colors easy to notice
\item A tie colored in \{color\} lies beside a handbag that has matching color accents, clearly showing their shared color
A \{color\} banana rests next to an apple that displays the same color as the banana’s peel.
\item A motorcycle painted \{color\} is parked next to a truck sharing the exact same color, making their colors clearly identical
A \{color\} goldfish is swimming near a turtle whose shell shows a similar color pattern.
\item A carrot with a \{color\} surface lies close to an orange that shares the same color, making the resemblance clear
\item A \{color\} baseball bat is resting against a skateboard that has the same color as the bat, showing a perfect match
An elephant painted \{color\} is standing near a giraffe whose colors closely resemble the elephant’s.
\item A \{color\} book is placed beside a clock that shares the same color, making it easy to see their matching appearance
\item A refrigerator painted \{color\} stands next to an oven that has been painted in the same color, clearly matching each other
\item A hat colored \{color\} rests on a jacket of identical color, making their shared shade obvious
\item A frisbee flying in \{color\} is near a kite that has the same color, making their similarity clear
A sports ball painted \{color\} lies next to a baseball glove that shares the same color.
\item A handbag colored \{color\} hangs near an umbrella with matching color tones, clearly showing they share the same color
A couch painted \{color\} is set beside a potted plant whose color scheme matches the couch’s perfectly.
\item A train painted \{color\} is passing near a taxi painted in the same color, making their matching colors easy to identify
\item A microwave colored \{color\} is placed on a counter next to a toaster with the same color, showing clear color correspondence
\item A football helmet painted \{color\} lies next to a boxing glove with matching color, making their similarity obvious
\item A table painted \{color\} is set near a candle that shares the same color, making their resemblance clear
A pair of jeans in \{color\} is folded next to pants that have the exact same color.
\item A car painted \{color\} is parked beside a minivan sharing the same color, making their colors clearly match
\item A tiger with \{color\} stripes is standing near a crocodile that has similar color tones, showing clear color resemblance
\item A toy terrier colored \{color\} sits close to a toy poodle that shares the same color, making their colors identical
\item A knife painted \{color\} is lying on a cutting board with matching color, showing a clear color match
\item A dog with \{color\} fur stands beside a horse that has the same color, making the shared color obvious
\item A suitcase colored \{color\} is placed next to a backpack of identical color, making their matching colors clear
\item A mug painted \{color\} is sitting next to a teapot with the same color, clearly matching each other
\item A car in \{color\} is parked beside a sports car of the same color, showing clear color similarity
\item An owl colored \{color\} is perched near a parrot sharing the same color, making the color match easy to see
\item A hair dryer painted \{color\} lies close to a remote control with matching color, clearly showing their similarity
\item A frisbee colored \{color\} is placed near a surfboard of the same color, showing clear color correspondence
\item An elephant painted \{color\} is standing next to a bear that has matching color, making the colors clearly match
\item A sweatshirt in \{color\} is folded beside a jacket that shares the same color, making their colors clearly identical
\item An apple colored \{color\} lies beside an orange that has the same color, making their similarity obvious
\item An airplane painted \{color\} is flying above a bus painted in the same color, showing a clear match
\item A skateboard colored \{color\} is lying near a pair of skis sharing the same color, making their colors match perfectly
\item A computer mouse colored \{color\} is lying near a cutting board that has the same color, making the color similarity clear
\item A banana with \{color\} peel is placed beside a mango that shares the same color, making their matching colors obvious
\item A sports ball colored \{color\} is lying near a football helmet with the same color, showing clear color similarity
\item A suitcase painted \{color\} is standing next to an umbrella of the same color, making the color match easy to see
\item A bear colored \{color\} is standing near a zebra that has matching color patterns, showing clear color resemblance
\item A remote control painted \{color\} is resting on a microwave that shares the same color, making the match clear
\item A dog with \{color\} fur is sitting next to a cat with identical color fur, making their colors obviously the same
\item An airplane painted \{color\} is flying above a truck painted the same color, making the color match obvious
\item A baseball bat colored \{color\} is lying near a frisbee of matching color, clearly showing their color similarity
\item A chair painted \{color\} is placed next to a table that has the same color, making the matching colors easy to see
\item A jacket colored \{color\} is hanging beside a coat of identical color, making their shared color obvious
\item A handbag painted \{color\} is resting on a backpack sharing the same color, clearly showing their color similarity
\item A car painted \{color\} is parked near a jeep painted the same color, making the matching colors obvious
\item A clock colored \{color\} is hanging near a vase of matching color, showing clear color correspondence
\item A crocodile colored \{color\} is swimming close to a shark with similar color, making their colors clearly alike
\item A toaster painted \{color\} is placed beside a refrigerator with matching color, showing their clear color match
\item A dog with \{color\} fur is standing beside a horse that has the same color, making the color similarity obvious
\item A sports ball painted \{color\} is lying near a golf ball with matching color, showing clear color resemblance
\item An umbrella colored \{color\} is hanging near a suitcase of the same color, making the colors clearly match
\item A chair painted \{color\} is placed next to a couch of identical color, showing their matching colors clearly
\item A baseball glove colored \{color\} is lying beside a baseball bat with matching color, making the color similarity obvious
\item A mango colored \{color\} is placed next to a papaya of the same color, showing clear color matching
\item A tie colored \{color\} is lying on a jacket with matching color, clearly showing their color similarity
\item A cat with \{color\} fur is sitting next to a dog with the same color fur, making the shared color obvious
\item A surfboard painted \{color\} is lying near a skateboard with identical color, making their colors clearly the same
\item A chair painted \{color\} is placed next to a desk with matching color, showing a clear color match
\item A microwave painted \{color\} is standing next to an oven of matching color, making the color similarity obvious
\item A baseball bat colored \{color\} is lying near a kite with the same color, clearly showing their matching colors
\item A bicycle painted \{color\} is parked beside a motorcycle sharing the same color, making their colors clearly alike
\item A parrot colored \{color\} is perched close to an owl with matching color, showing their shared color clearly
\item A dog with \{color\} fur is sitting beside a teddy bear of the same color, making the color match obvious
\item A football helmet painted \{color\} is lying near a snowboard with matching color, showing clear color similarity
\item An apple colored \{color\} is placed next to a guava of the same color, making the matching colors easy to see
\item A chair painted \{color\} is placed near a bookcase with matching color, making their colors clearly match
\item A suitcase colored \{color\} is resting next to a handbag of identical color, showing clear color correspondence
\item An orange colored \{color\} is lying near a carrot with the same color, making their colors clearly alike
\item A refrigerator painted \{color\} is standing next to a toaster with matching color, showing clear color similarity
\item A horse colored \{color\} is standing beside a sheep sharing the same color, making the color match obvious
\item A sports ball painted \{color\} is lying near a football helmet with matching color, showing clear color resemblance
\item A handbag colored \{color\} is placed next to a coat of the same color, making the colors clearly the same
\item A baseball bat colored \{color\} is lying near a sports ball of matching color, showing clear color similarity
\item A car painted \{color\} is parked beside a taxi painted the same color, making the colors clearly match
\item A clock colored \{color\} is hanging near a vase of matching color, showing clear color correspondence
\item A tie colored \{color\} is lying near a pair of pants with the same color, making the matching colors obvious
\item A dog with \{color\} fur is sitting beside a cat with identical color fur, showing their matching colors clearly
\item A baseball glove colored \{color\} is lying near a baseball bat with matching color, making their color similarity clear
\item A motorcycle painted \{color\} is parked next to a truck sharing the same color, making the color match obvious
\item An apple colored \{color\} is lying beside an orange of the same color, showing their matching colors clearly
\item A sweatshirt colored \{color\} is lying beside a pair of jeans with matching color, making their colors clearly the same
\item A remote painted \{color\} is resting on a microwave of the same color, making their color similarity clear
\item An umbrella colored \{color\} is hanging near a backpack with matching color, showing clear color correspondence
\item A banana colored \{color\} is placed next to a mango of the same color, making their colors clearly alike
\item A knife colored \{color\} is lying on a cutting board with matching color, showing clear color similarity
\item A dog with \{color\} fur is standing near a horse with identical color, making the matching colors obvious
\item A tie colored \{color\} is lying over a jacket of the same color, making their colors clearly the same
\item A surfboard colored \{color\} is lying near a skateboard sharing the same color, showing clear color resemblance
\item A handbag painted \{color\} is resting on a suitcase with matching color, making the color match obvious
\item A chair colored \{color\} is placed next to a couch of identical color, showing clear color similarity
\item A football helmet painted \{color\} is lying near a boxing glove with the same color, making their colors clearly the same
\item A tiger colored \{color\} is standing close to a crocodile sharing the same color, showing clear color correspondence
\end{enumerate}

\begin{figure*}[t]
\centering
\includegraphics[width=\linewidth]{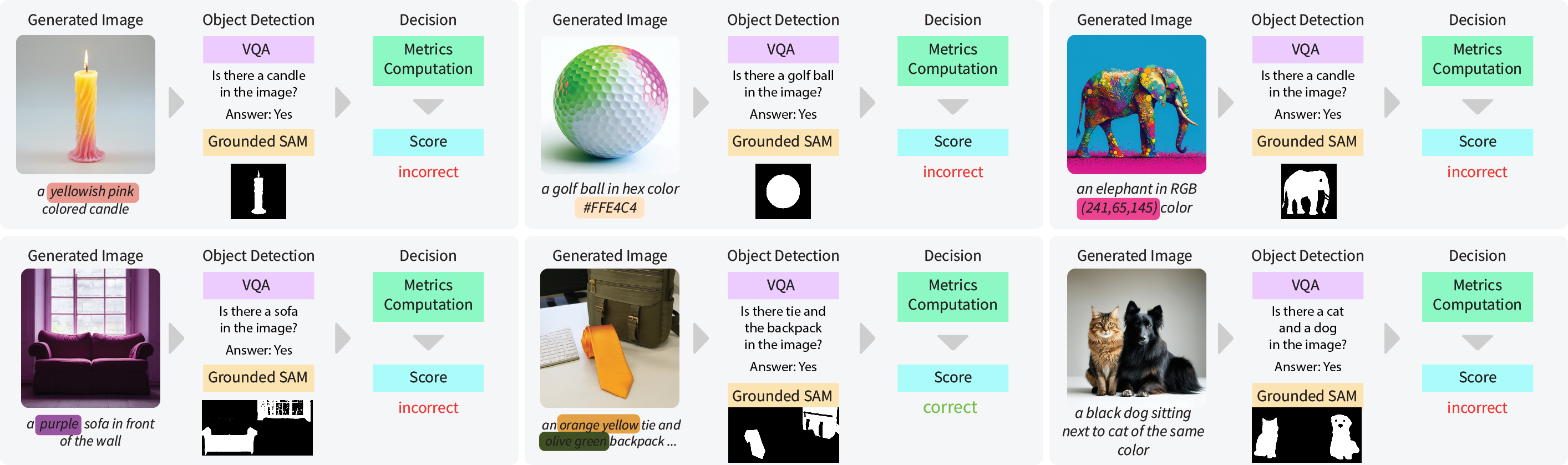}
\caption{Qualitative Examples of the Color Generations of T2I models for different prompt settings.}
\label{fig:results}
\end{figure*}

\begin{table*}[t]
\centering
\resizebox{\textwidth}{!}{
\begin{tabular}{l|c|c|c|c|c|c|c|c|c|c}
\hline
\multicolumn{11}{c}{\textbf{CSS3/X11 Color Space (2,058 samples, 6 questions)}} \\
\hline
\textbf{Model} & \textbf{Overall} & \textbf{Open-Ended} & \textbf{MCQ} & \textbf{Binary} & \textbf{Q1: Name} & \textbf{Q2: MCQ} & \textbf{Q3: Hex} & \textbf{Q4: Name Ver.} & \textbf{Q5: RGB Ver.} & \textbf{Q6: Hex Ver.} \\
\hline
Janus & 26.56 & 5.03 & 12.20 & 52.44 & 7.34 & 12.20 & 2.72 & 57.43 & 50.00 & 49.90 \\
Janus Pro & 28.64 & 6.63 & 19.44 & 50.95 & 7.48 & 19.44 & 5.78 & 55.00 & 47.18 & 50.68 \\
BLIP3o & 31.41 & 12.46 & 24.73 & 57.08 & 9.96 & 24.73 & 14.38 & 60.50 & 57.05 & 53.69 \\
Deepseek & 28.72 & 11.35 & 18.85 & 55.85 & 9.43 & 18.85 & 13.27 & 56.41 & 53.60 & 57.53 \\
Qwen2-VL & 29.13 & 9.36 & 23.23 & 54.81 & 8.94 & 23.23 & 9.77 & 55.20 & 53.21 & 56.03 \\
Instruct-VL & 27.96 & 7.19 & 20.55 & 56.17 & 8.94 & 20.55 & 5.44 & 56.75 & 55.00 & 56.75 \\
mPLUG-Owl3 & 26.76 & 7.24 & 17.93 & 55.00 & 6.95 & 17.93 & 7.53 & 56.41 & 54.37 & 54.23 \\

\end{tabular}
}
\resizebox{\textwidth}{!}{
\begin{tabular}{l|c|c|c|c|c|c|c|c}
\hline
\multicolumn{9}{c}{\textbf{IBCC L2 Color Space (406 samples, 4 questions)}} \\
\hline
\textbf{Model} & \textbf{Overall} & \textbf{Open-Ended} & \textbf{MCQ} & \textbf{Binary} & \textbf{Q1: Name} & \textbf{Q2: MCQ} & \textbf{Q3: Name Ver.} & \textbf{Q4: RGB Ver.} \\
\hline
Janus & 37.68 & 25.86 & 33.99 & 55.79 & 25.86 & 33.99 & 61.58 & 50.00 \\
Janus Pro & 44.02 & 26.60 & 43.60 & 55.91 & 26.60 & 43.60 & 64.53 & 47.29 \\
BLIP3o & 45.13 & 25.12 & 45.81 & 64.41 & 25.12 & 45.81 & 72.41 & 56.40 \\
Deepseek & 45.50 & 27.34 & 45.32 & 63.30 & 27.34 & 45.32 & 72.91 & 53.69 \\
Qwen2-VL & 45.16 & 24.63 & 43.35 & 63.67 & 24.63 & 43.35 & 69.46 & 57.88 \\
Instruct-VL & 44.91 & 26.60 & 45.57 & 62.56 & 26.60 & 45.57 & 72.41 & 52.71 \\
mPLUG-Owl3 & 44.95 & 24.14 & 42.12 & 62.57 & 24.14 & 42.12 & 72.91 & 52.22 \\
\hline
\end{tabular}
}
\caption{Comprehensive Performance Analysis of VLM-based VQA on ISCC NBS Level 2 and CSS3/X11 colors.}
\label{tab:comprehensive_performance}
\end{table*}

\section{Limitations of VQA-based Assessment Methods}
Table~\ref{tab:comprehensive_performance} presents a detailed evaluation of three visual-language models—Janus, Janus Pro, and mPLUG-large—on color understanding tasks grounded in two standard color sets: CSS3/X11 and ISCC-NBS Level 2. The goal of this benchmark is to probe how well VLMs can interpret, reason about, and verify colors in images using natural language.

We design a set of structured visual question answering (VQA) tasks, covering three reasoning modes:
(1) Open-Ended Question, where the model must produce a color name or code in free-form;
(2) Multiple Choice Question, where it must choose the correct answer from a list; and
(3) Binary Question, where it must answer "Yes" or "No" to a color-specific query.

\begin{figure*}[t]
    \centering
    \includegraphics[width=\linewidth]{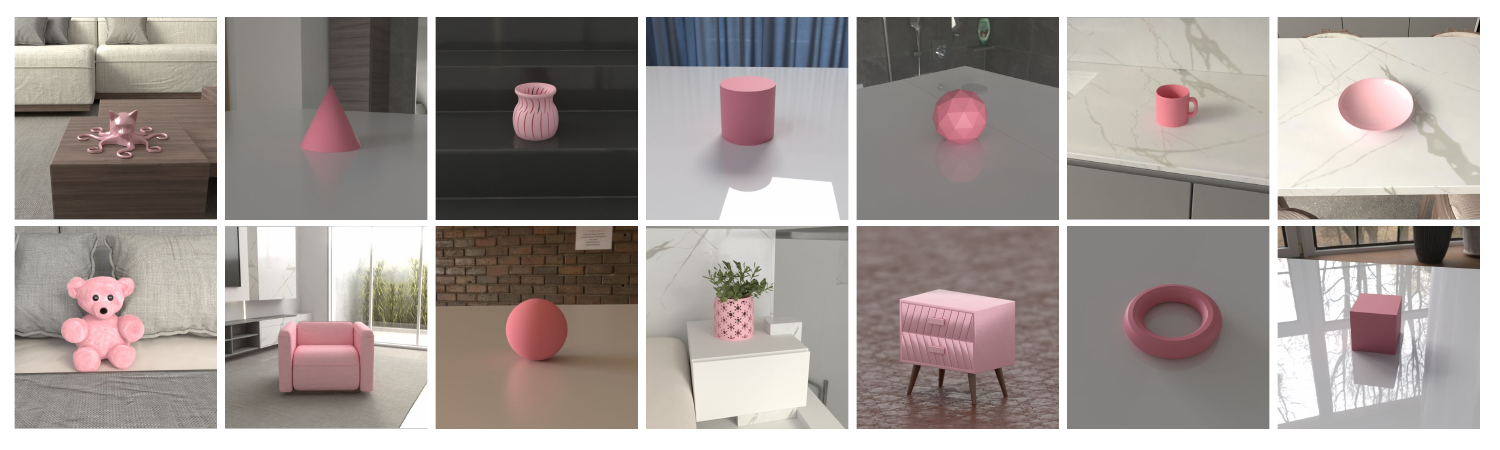}
    \caption{Set of 14 objects rendered in blender in ISCC-NBS Level 2 and CSS3/X11 colors for VQA evaluation.}
    \label{fig:blend_shapes}
\end{figure*}

Each model is tested on a set of image-question pairs derived from real generations using known target colors. we render the 14 objects in the CSS3/X11 and ISCC-NBS L2 colors in blender, as shown in Figure.~\ref{fig:blend_shapes}.

We evaluate six question types under the CSS3/X11 taxonomy (2058 samples), and four question types for ISCC-NBS Level 2 (406 samples), listed below:

\vspace{0.3em}
\noindent\textbf{CSS3/X11 Questions:}
\begin{itemize}
    \item Q1 (Generative): What is the CSS3/X11 color name of the \{object\} in the given image?
    \item Q2 (Discriminative): Given the list of CSS3/X11 color names [...], what is the color name of the \{object\} in the image?
    \item Q3 (Generative): What is the CSS3/X11 hex color code of the \{object\} in the image?
    \item Q4 (Verification): Is the color name of the \{object\} \{color\} in the image?
    \item Q5 (Verification): Is the RGB color of the \{object\} \{r,g,b\} in the image?
    \item Q6 (Verification): Is the hex color code of the \{object\} \{hex code\} in the image?
\end{itemize}

\vspace{0.3em}
\noindent\textbf{ISCC-NBS Level 2 Questions:}
\begin{itemize}
    \item Q1 (Generative): What is the IBCC Level 2 color name of the \{object\} in the image?
    \item Q2 (Discriminative): Given the list of IBCC Level 2 color names [...], what is the color name of the \{object\}?
    \item Q3 (Verification): Is the color name of the \{object\} \{color\} in the image?
    \item Q4 (Verification): Is the RGB color of the \{object\} \{r,g,b\} in the image?
\end{itemize}

From the results presented in Table~\ref{tab:comprehensive_performance}, we observe that binary verification tasks consistently yield the highest accuracy, while open-ended generative tasks are particularly weak across all models and color spaces. For instance, in the CSS3/X11 benchmark, open-ended scores (Q1 and Q3) remain below 8\% across all models, with the best-performing model (BLIP3o) achieving only 12.46\% in open-ended tasks overall. This suggests that models struggle to produce the correct color name or hex code when not explicitly prompted with options.

In contrast, binary verification tasks (Q4–Q6) see much higher accuracy—often above 50\%—indicating that VLMs are better at recognizing and verifying predefined information than generating it. MCQ tasks (Q2) perform moderately well, achieving up to 24.73\% (BLIP3o), but still depend on the presence of semantically close distractors and color naming consistency.

Similar trends appear in the IBCC L2 evaluation. Although the smaller color set improves overall performance, open-ended results remain limited (e.g., Deepseek: 27.34\%), and binary verification continues to dominate (e.g., achieving up to 64.41\% in BLIP3o and 63.67\% in Qwen2-VL).

These results highlight a critical limitation that VLMs are not reliable for evaluating color fidelity. First, their low open-ended accuracy indicates poor internal representation of precise color semantics despite correct visual grounding. Second, their strong binary verification performance may result from learning patterns in color datasets rather than actual understanding. Third, performance varies significantly across color taxonomies and question types, exposing instability in color reasoning.

Ultimately, while VLMs can aid in approximate visual understanding, they are not suitable as color evaluation agents in benchmarks requiring accurate color reproduction. This motivates the need for dedicated, metric-based evaluation pipeline rather than relying on VQA responses for color assessment.

\section{Category-wise Qualitative Analysis of T2I models}

\begin{table*}[t]
\centering
\tiny
\begin{tabular}{|c|l|c|c|c|c|c|c|c|}
\hline
\textbf{Model} & \textbf{Color Type} & \textbf{Clothes\_Acc} & \textbf{Vehicle} & \textbf{Furniture\_House} & \textbf{Tools\_Misc} & \textbf{Sports\_Toys} & \textbf{Animals} & \textbf{Fruit\_Veg} \\
\hline
Flux  & css     & 39.5554 & 39.4843 & 37.4237 & 35.9550 & 34.9350 & 30.5794 & 21.2568 \\
Flux  & l2      & 60.4044 & 56.9160 & 51.2754 & 50.3166 & 49.2966 & 41.9628 & 37.4442 \\
Flux  & l3      & 51.9078 & 49.9698 & 45.2676 & 43.8498 & 43.2990 & 38.4948 & 29.6718 \\
\hline
\rowcolor{lightgray} Flux  & overall & 50.6226 & 48.7899 & 44.6556 & 43.3737 & 31.0616 & 37.0124 & 29.4576 \\
\hline
PixArt-Alpha &  css     & 41.3814 & 50.4084 & 44.8596 & 45.8796 & 44.4516 & 40.6266 & 24.2250 \\
PixArt-Alpha &  l2      & 62.1486 & 56.4468 & 64.3722 & 56.7834 & 58.2930 & 55.8348 & 38.8824 \\
PixArt-Alpha &  l3      & 56.5794 & 46.6446 & 55.3962 & 48.7866 & 52.5402 & 45.9816 & 31.2018 \\
\hline
\rowcolor{lightgray} PixArt-Alpha &  overall & 53.3700 & 51.1666 & 54.8760 & 50.4832 & 51.7616 & 47.4810 & 31.4364 \\
\hline
PixArt-Sigma &  css     & 37.8318 & 42.7380 & 42.4932 & 42.6156 & 44.5128 & 37.9338 & 29.2332 \\
PixArt-Sigma &  l2      & 58.7520 & 51.9588 & 62.5668 & 57.0048 & 58.8744 & 47.7972 & 46.9098 \\
PixArt-Sigma &  l3      & 55.6818 & 43.2480 & 53.6724 & 46.8078 & 52.6320 & 38.0256 & 34.1086 \\
\hline
\rowcolor{lightgray} PixArt-Sigma &  overall & 50.7552 & 45.9816 & 52.9108 & 48.8036 & 52.0064 & 41.2522 & 36.7472 \\
\hline
Sana &  css     & 46.1550 & 50.5818 & 44.2068 & 47.5728 & 51.4794 & 48.3378 & 23.3376 \\
Sana &  l2      & 62.3730 & 50.7960 & 58.4154 & 57.4974 & 60.5370 & 58.1298 & 40.2186 \\
Sana &  l3      & 58.1706 & 45.6144 & 53.1012 & 49.2354 & 54.6312 & 50.0514 & 22.0422 \\
\hline
\rowcolor{lightgray} Sana &  overall & 55.5662 & 48.9974 & 51.9078 & 51.4352 & 55.5492 & 52.1730 & 28.5328 \\
\hline
SD 3.5 &  css     & 42.7176 & 45.4920 & 41.8506 & 44.5536 & 38.0460 & 38.0154 & 22.3788 \\
SD 3.5 &  l2      & 64.0458 & 59.2926 & 61.9848 & 61.3530 & 56.4468 & 52.0812 & 39.0660 \\
SD 3.5 &  l3      & 62.1384 & 52.9686 & 56.7630 & 51.6936 & 53.1624 & 48.9804 & 28.4478 \\
\hline
\rowcolor{lightgray} SD 3.5 &  overall & 56.3006 & 52.5844 & 53.5466 & 52.5334 & 49.2184 & 46.3590 & 30.0642 \\
\hline
SD 3 &  css     & 42.5034 & 43.7070 & 41.8000 & 42.7788 & 41.4936 & 40.6062 & 25.0716 \\
SD 3 &  l2      & 55.4370 & 48.6438 & 56.5386 & 51.0306 & 44.8596 & 49.7964 & 37.1178 \\
SD 3 &  l3      & 54.0030 & 48.9804 & 52.9980 & 52.8360 & 42.0444 & 47.6646 & 32.0586 \\
\hline
\rowcolor{lightgray} SD 3 &  overall & 50.6430 & 47.1104 & 50.4594 & 48.8818 & 42.7992 & 45.9924 & 31.4160 \\
\hline
Janus-Pro &  css     & 24.0822 & 25.4490 & 28.0602 & 25.4490 & 25.2246 & 13.1886 & 7.8642 \\
Janus-Pro &  l2      & 46.4100 & 35.6898 & 41.8914 & 34.4862 & 39.2600 & 26.1834 & 19.2576 \\
Janus-Pro &  l3      & 42.8400 & 31.0794 & 39.9330 & 32.3442 & 35.6490 & 24.3474 & 16.1772 \\
\hline
\rowcolor{lightgray} Janus-Pro &  overall & 37.7774 & 30.7394 & 36.6282 & 30.7600 & 33.3778 & 21.2398 & 14.4330 \\
\hline
OmniGen2 &  css     & 35.4246 & 34.0884 & 36.3324 & 39.6270 & 41.3508 & 27.2034 & 23.6334 \\
OmniGen2 &  l2      & 60.5472 & 46.1856 & 60.7308 & 54.3660 & 55.8144 & 41.3814 & 38.0766 \\
OmniGen2 &  l3      & 48.3886 & 38.8416 & 49.0314 & 45.0534 & 47.0120 & 34.8024 & 28.4580 \\
\hline
\rowcolor{lightgray} OmniGen2 &  overall & 48.1168 & 39.7052 & 48.6982 & 46.3488 & 48.0488 & 34.4624 & 30.0560 \\
\hline
BLIP3o &  css     & 34.1496 & 35.1696 & 33.6192 & 43.8294 & 42.1464 & 24.6228 & 28.8048 \\
BLIP3o &  l2      & 47.9706 & 42.2076 & 45.9204 & 54.5496 & 52.3464 & 33.8130 & 42.4218 \\
BLIP3o &  l3      & 46.6140 & 37.3932 & 37.5462 & 49.0518 & 43.5540 & 30.0180 & 34.8942 \\
\hline
\rowcolor{lightgray} BLIP3o &  overall & 42.9114 & 38.2568 & 39.0286 & 49.1436 & 45.9956 & 29.4678 & 35.3736 \\
\hline
\end{tabular}
\caption{Per-category performance across all models and scoring types (css, l2, l3, overall).}
\label{tab:per_cat_acc}
\end{table*}

Table~\ref{tab:per_cat_acc} reveals strong evidence of model bias and entanglement between object semantics and color generation. Across all the employed models, performance varies significantly depending on both the object category and the chosen color taxonomy. For example, the models perform better on categories like Clothing, Furniture, and Vehicles, while they struggle on categories like Fruits and Vegetables and Animals. This discrepancy indicates that models are not disentangling color prompts from natural objects training priors. Instead, they tend to reproduce default or canonical colors seen during training (e.g., red apples, yellow bananas), even when explicitly instructed otherwise. Some examples are shown in the Figure~\ref{fig:results}.

Furthermore, the models perform markedly better with coarse-grained color taxonomies like ISCC-NBS Level 2, but struggle with the finer granularity of CSS3 and Level 3, indicating a lack of standard color understanding. The combination of these effects highlights that the current T2I models often conflate object identity with memorized color distributions, failing to generalize to atypical or prompt-specified colors. This entanglement points to a systemic bias in training data and architecture, which limits their usefulness for tasks requiring precise and independent control over object appearance attributes—such as color.

\begin{figure*}
    \centering
    \includegraphics[width=\linewidth]{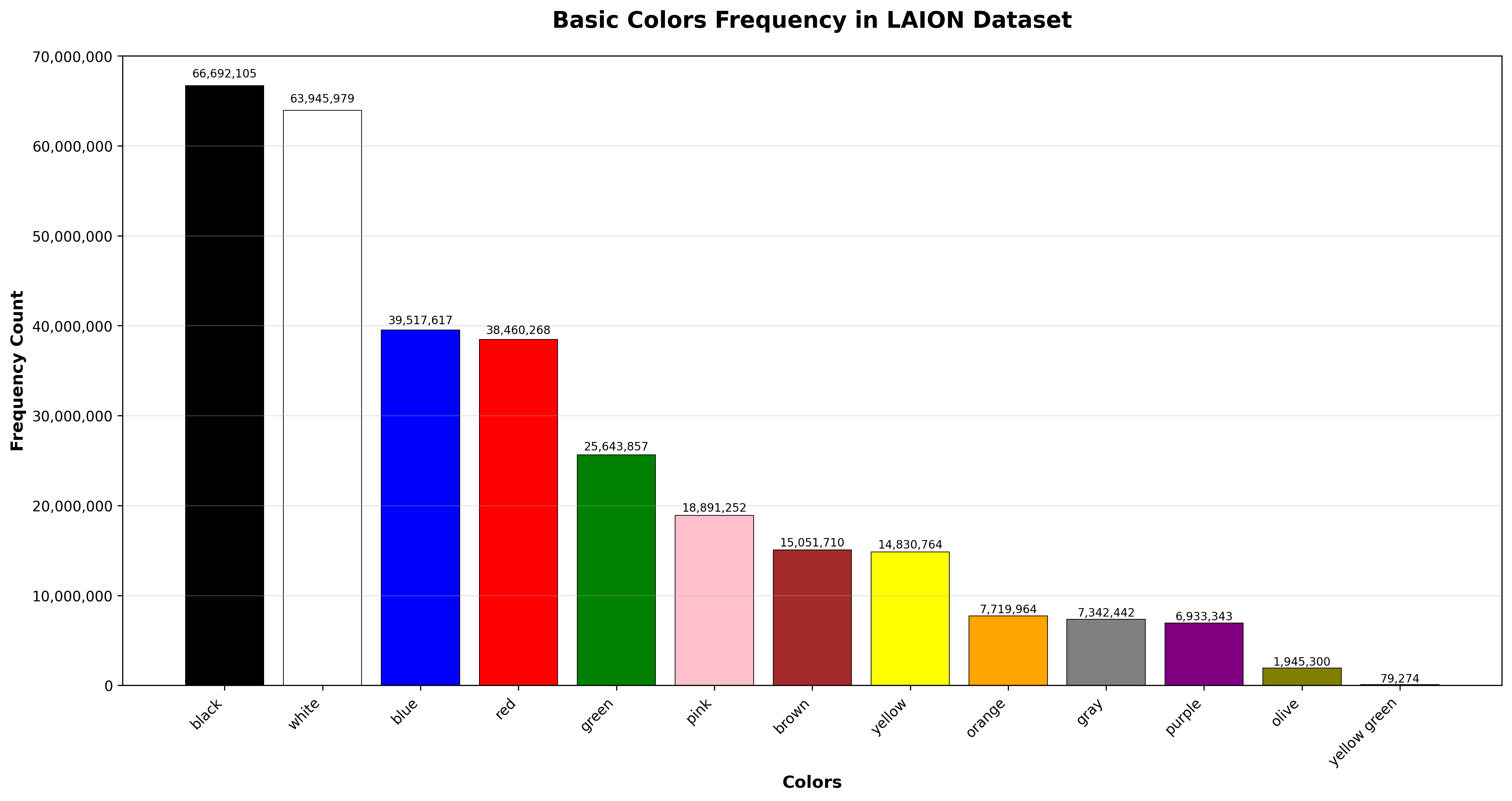}
    \caption{Frequency Analysis of Basic Colors in LAION-2B text prompts.}
    \label{fig:basic_colors}
\end{figure*}

\begin{figure*}
    \centering
    \includegraphics[width=\linewidth]{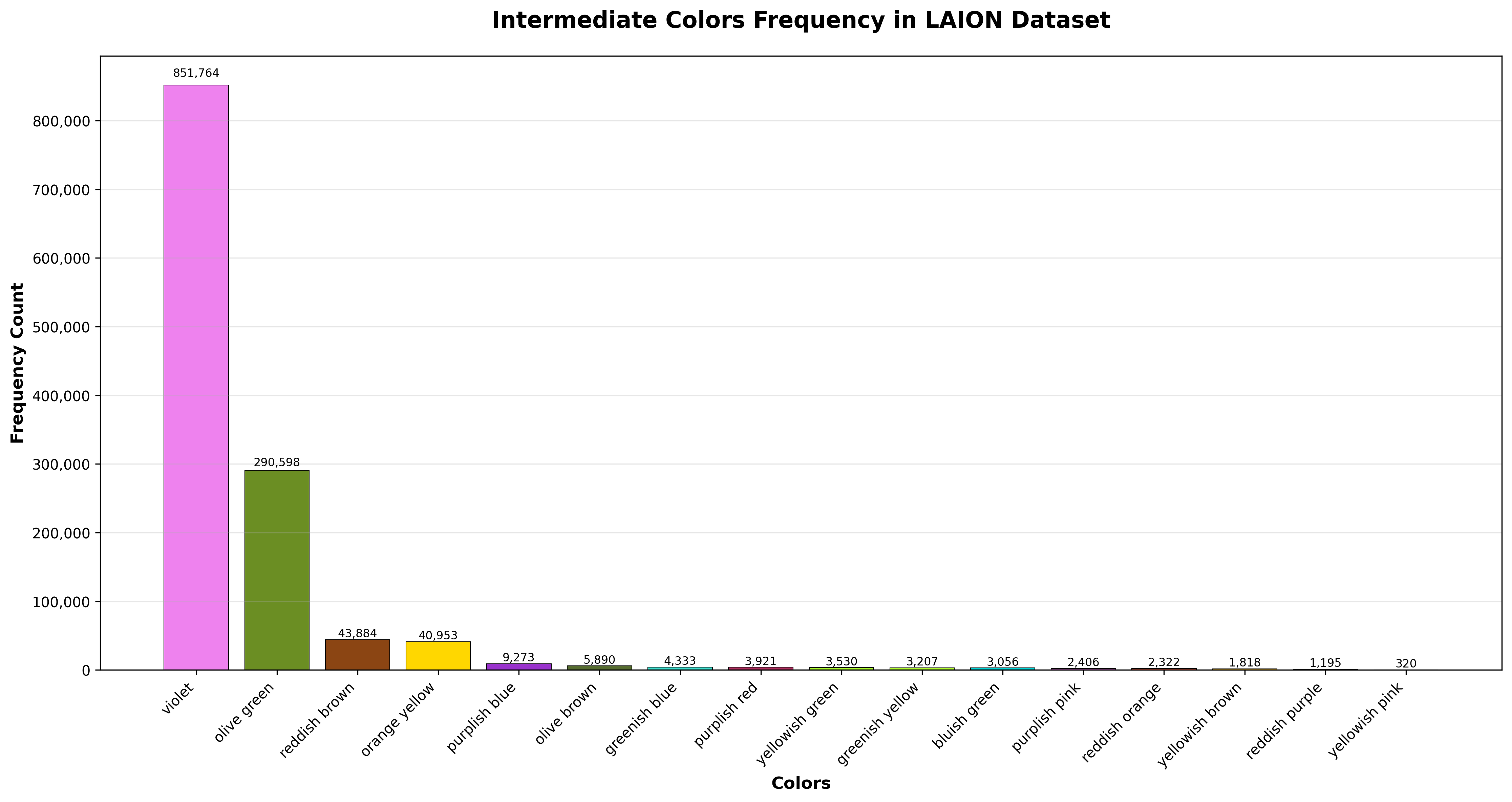}
    \caption{Frequency Analysis of Intermediate Colors in LAION-2B text prompts.}
    \label{fig:intermediate_colors}
\end{figure*}

\begin{figure*}[t]
    \centering
    \includegraphics[width=\linewidth]{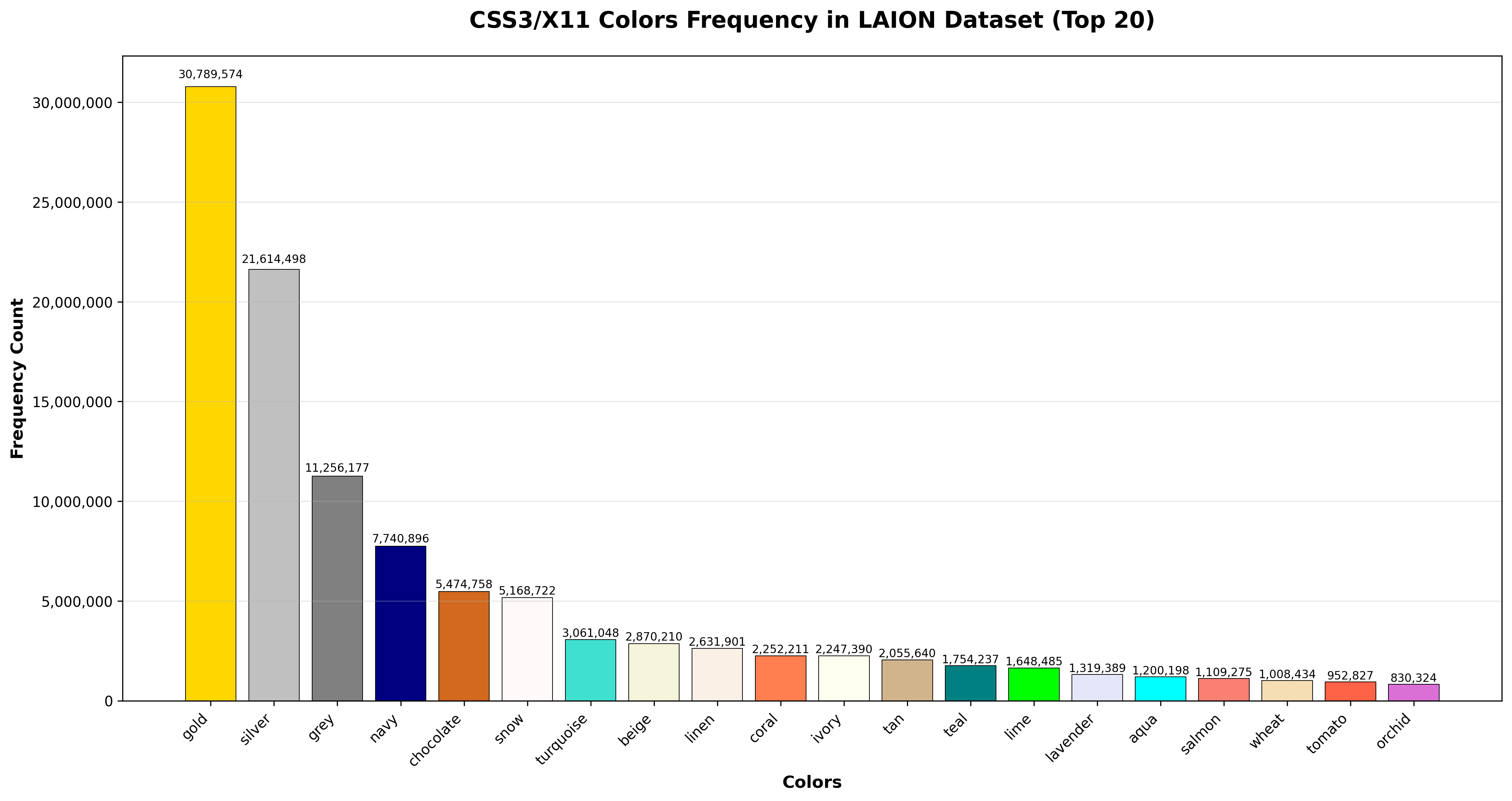}
    \caption{Frequency Analysis of CSS3/X11 Colors in LAION-2B text prompts.}
    \label{fig:css3x11_colors}
\end{figure*}

\begin{figure*}[t]
    \centering
    \includegraphics[width=\linewidth]{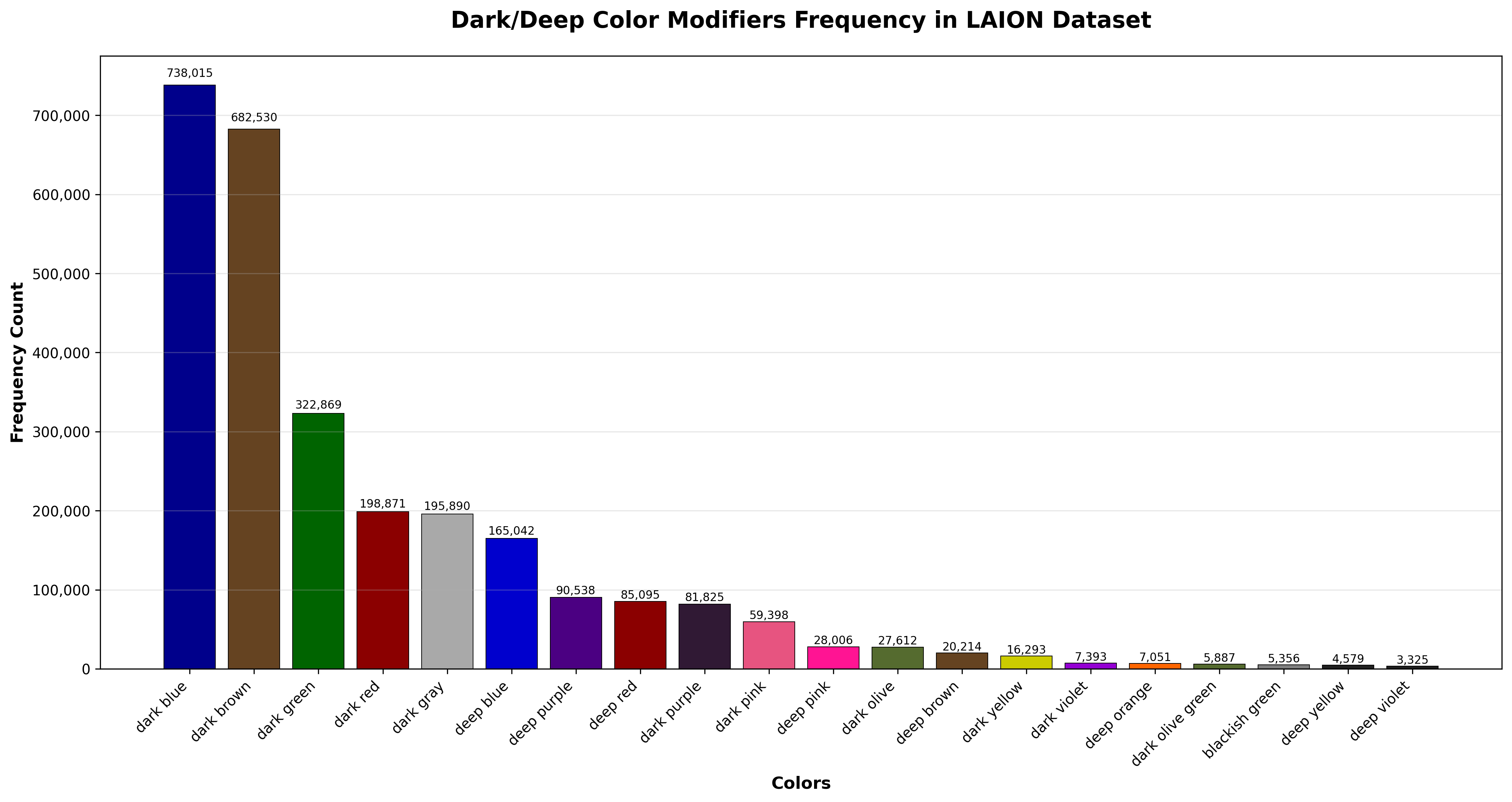}
    \caption{Frequency Analysis of Dark Color Modifiers in LAION-2B text prompts.}
    \label{fig:dark_modifiers}
\end{figure*}

\begin{figure*}[!t]
    \centering
    \includegraphics[width=\linewidth]{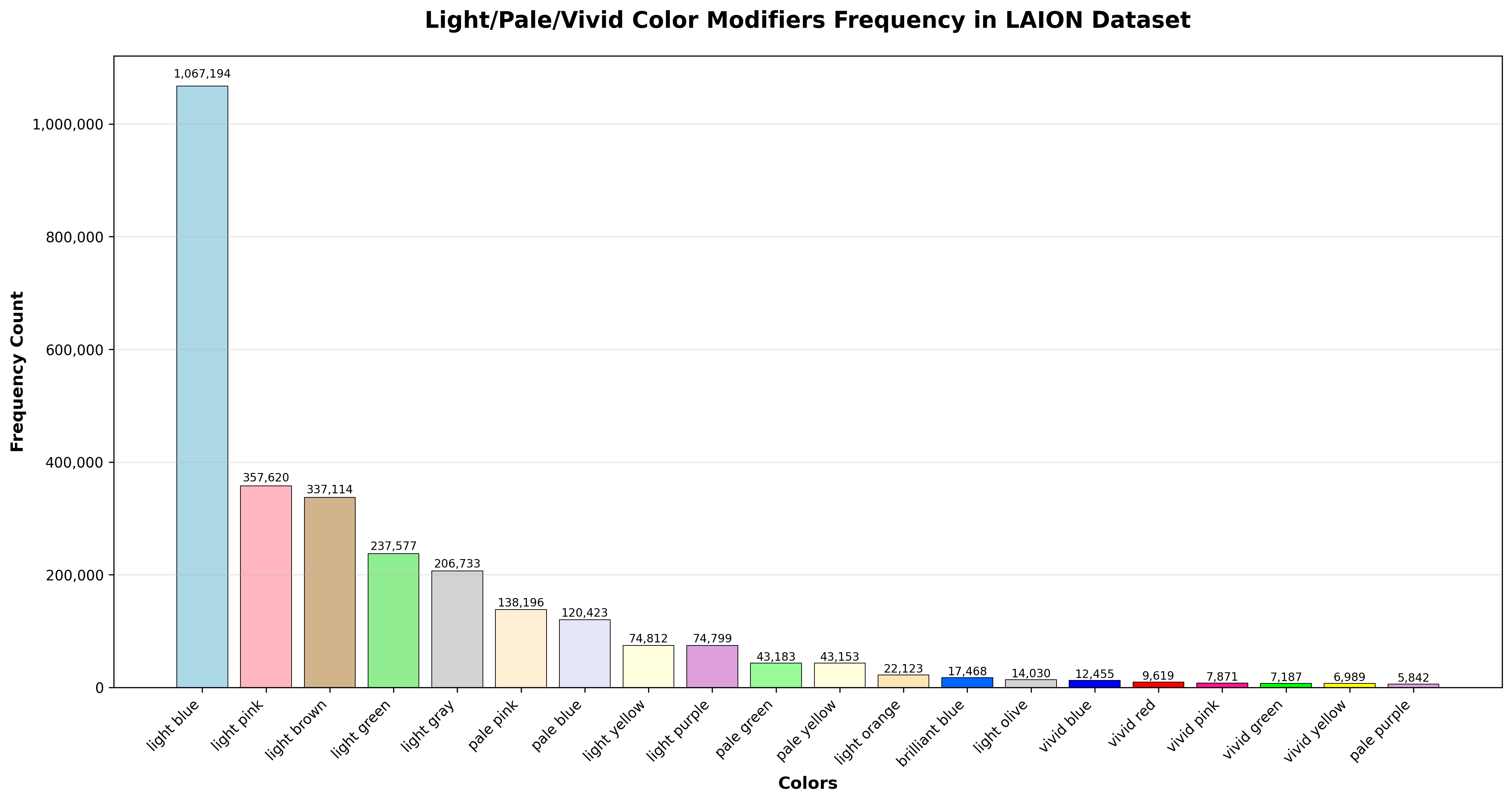}
    \caption{Frequency Analysis of Light Color Modifiers in LAION-2B text prompts.}
    \label{fig:light_colors}
\end{figure*}

\end{document}